\documentclass[runningheads]{llncs}

\usepackage[utf8]{inputenc}
\usepackage[T1]{fontenc}
\usepackage{booktabs}
\usepackage[table]{xcolor}
\usepackage{longtable}
\usepackage{array}
\usepackage{enumitem}
\usepackage{microtype}
\usepackage{parskip}
\usepackage{amssymb}
\usepackage{graphicx}
\usepackage{colortbl}
\usepackage{multirow}
\usepackage{subcaption}
\usepackage{wrapfig}
\definecolor{searow}{gray}{0.92}

\definecolor{highlightgreen}{RGB}{198,239,206}
\definecolor{highlightyellow}{RGB}{255,235,156}

\definecolor{bestrow}{RGB}{220,230,255}

\usepackage{accv}

\usepackage{accvabbrv}

\usepackage{graphicx}
\usepackage{booktabs}

\usepackage[accsupp]{axessibility}  

\usepackage{hyperref}

\usepackage{orcidlink}

\begin{document}

\title{SEA-CLIP-Tiny: Efficient Multilingual Text-Vision Embedding for Southeast Asian Languages} 

\titlerunning{SEA-CLIP-Tiny}


\author{
    Puja Ahmad Habibi\inst{1}\thanks{Equal contribution.} \and
    Faiz Assabil Firdaus\inst{1,2}\protect\footnotemark[1] \and
    Ashvanth S\inst{1,3,4}\protect\footnotemark[1] \and
    Ekapol Chuangsuwanich\inst{1,5} \and
    Pume Tuchinda\inst{1,6,7} \and
    Peerat Limkonchotiwat\inst{1,7}
}

\authorrunning{P.A. Habibi et al.}

\institute{
    SEACrowd \and
    University of Indonesia \and
    Cohere Labs Community \and
    Technical University of Denmark \and
    Department of Computer Engineering, Faculty of Engineering, Chulalongkorn University \and
    Vidyasirimedhi Institute of Science and Technology \and
    AI Singapore\\
    \email{peerat@aisingapore.org}
}
\maketitle

\begin{abstract} 
Multilingual text-vision embedding models are essential for cross-lingual image-text retrieval, but Southeast Asian languages remain poorly supported due to the region's linguistic diversity and limited data and computing resources. In this paper, we introduce \textbf{SEA-CLIP-Tiny}, a compact multilingual text-vision embedding model for Southeast Asia with fewer than 50M parameters. Our model adapts a CLIP-KD-style framework to Southeast Asian multilingual settings through regional data curation and multilingual teacher guidance. Experiments across seven Southeast Asian languages show that \textbf{SEA-CLIP-Tiny} achieves the strongest average retrieval performance among the evaluated student models, reaching 12.9\%, 31.5\%, and 42.2\% at R@1, R@5, and R@10, respectively. Compared with MobileCLIP2, it improves average R@10 by 12.1 points while using 38.4\% fewer parameters and lower measured CPU latency. These results highlight the importance of region-aware training for efficient multilingual text-vision models in Southeast Asia. We release the \textbf{SEA-CLIP-Tiny} model weights and datasets (\url{https://huggingface.co/collections/fassabilf/sea-clip-tiny-accv-2026}) and the training, evaluation, and preprocessing code (\url{https://github.com/fassabilf/sea-clip-tiny}).

  \keywords{Multilingual CLIP \and Low-Resource Languages \and Knowledge Distillation}
\end{abstract}

\section{Introduction}
\label{sec:introduction}
Multilingual text-vision embedding models aim to align images and texts from multiple languages in a shared representation space, following the contrastive image-text pretraining paradigm popularized by CLIP~\cite{radford2021learning}. 
By enabling language-agnostic visual understanding, such models support cross-lingual image retrieval, zero-shot classification, multimodal search, and downstream vision-language applications for users beyond English-speaking communities~\cite{10656429,Vayani_2025_CVPR}.
The ability to retrieve and reason over visual content using natural language is essential for building inclusive multimodal systems. 
In particular, multilingual text-vision embedding models allow users to query images in their native languages without relying on translation systems, which may introduce errors or lose cultural specific meaning. 
Therefore, they are an important foundation for practical applications such as multilingual e-commerce search, education, digital archives, and assistive technologies~\cite{chen-etal-2023-mclip,nllb_clip}.
Recent advances in multilingual vision-language modeling have shown promising progress through large-scale models such as SigLIP2~\cite{siglip2}, Qwen-VL~\cite{qwen_vl}, and other multilingual CLIP-style~\cite{chen-etal-2023-mclip,xu2023metaclip,chuang2025metaclip2} or vision-language models~\cite{qwen_vl}. These models demonstrate strong capabilities across a range of high-resource languages and general multimodal tasks. However, despite their broad multilingual coverage, the support of SEA languages is not demonstrated or supported compared to English or other high-resource languages.

\begin{figure}[t]

    \centering

    \begin{subfigure}{0.42\linewidth}
        \centering
        \includegraphics[width=\linewidth]{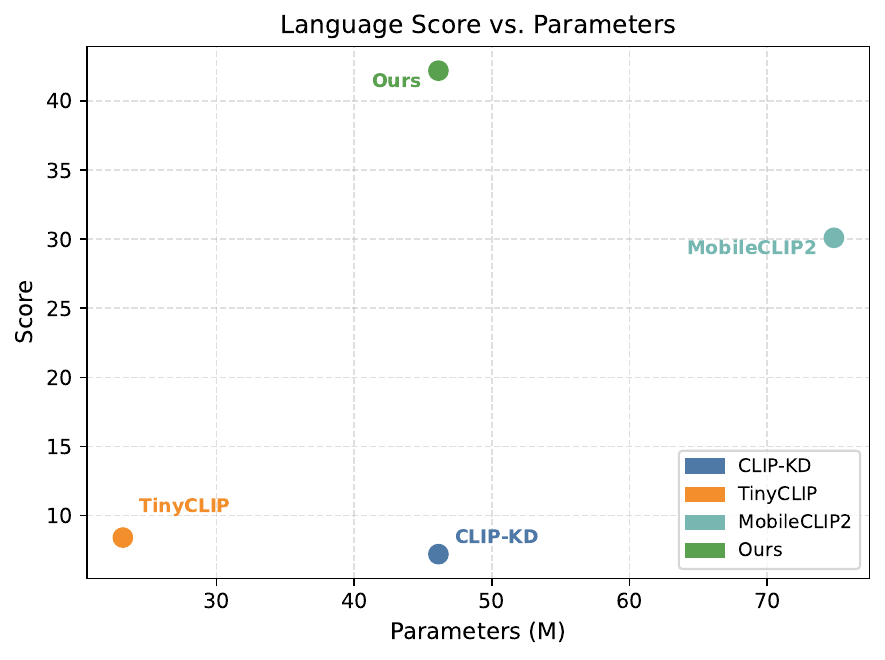}
        \caption{Language Score vs. Parameters.}
        \label{fig:plot-score-1}
    \end{subfigure}
    \hfill
    \begin{subfigure}{0.42\linewidth}
        \centering
        \includegraphics[width=\linewidth]{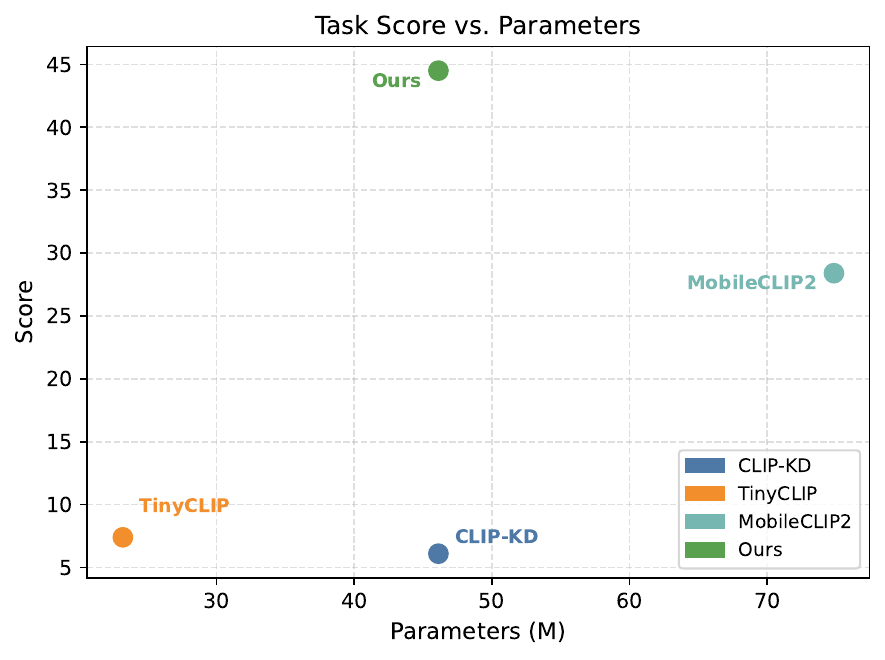}
        \caption{Task Score vs. Parameters.}
        \label{fig:plot-score-2}
    \end{subfigure}
    \vspace{-2mm}
    \caption{Performance-size comparison of compact CLIP models on Southeast Asian image-text retrieval. (a) Language-average R@10 is computed from mean(I2T, T2I) retrieval scores over the available retrieval benchmarks for each of the seven SEA languages. (b) Task-average R@10 is the mean retrieval R@10 over XM3600, Flickr30k-200, and XTD-200.}
    \label{fig:score-plots}
    \vspace{-5mm}
\end{figure}

\begin{wrapfigure}{r}{0.40\textwidth}
    \vspace{-8mm}
    \centering
    \includegraphics[width=0.40\textwidth]{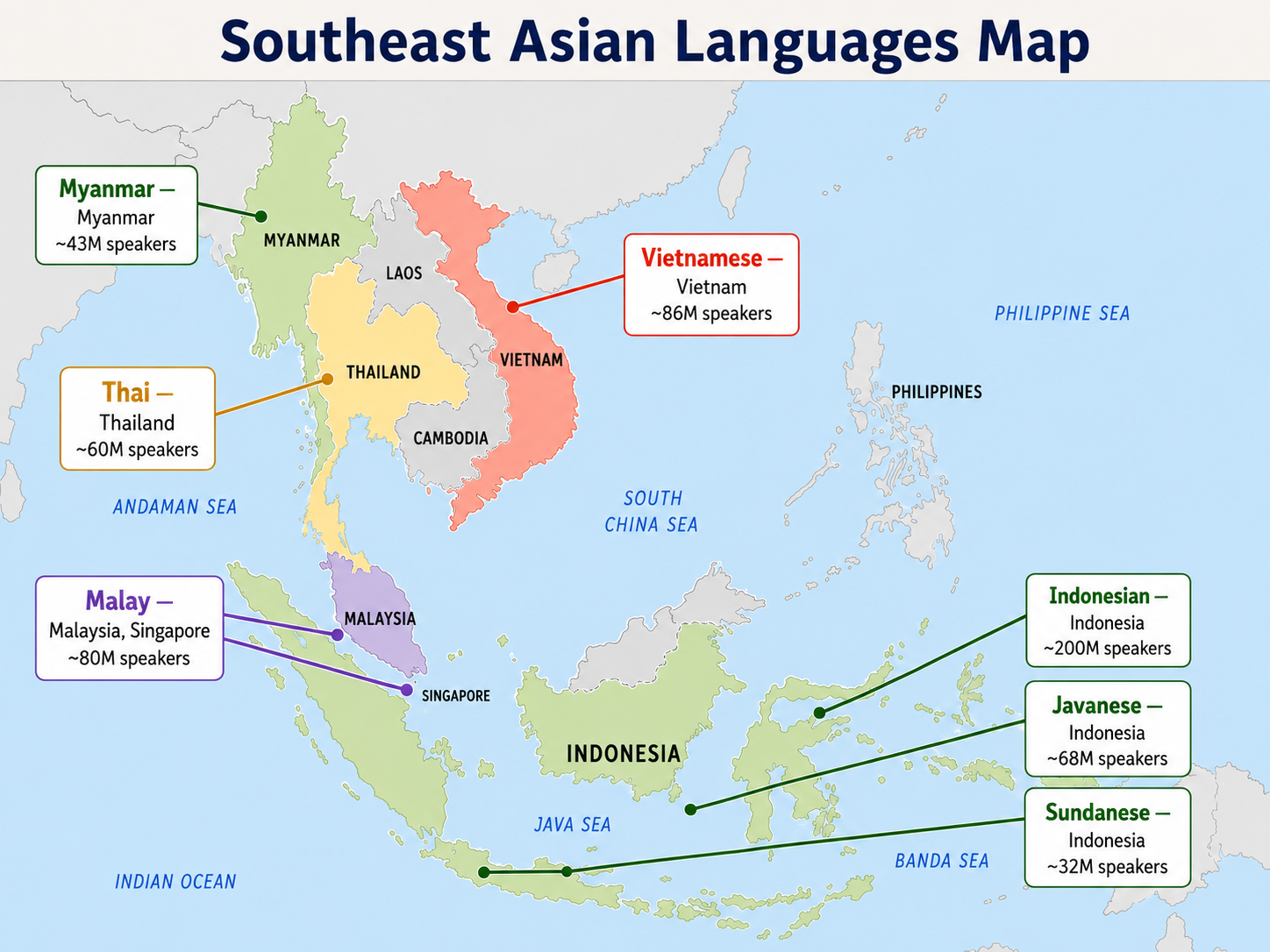}
    \vspace{-5mm}
    \caption{SEA languages and countries covered in this work, amounting to 569M speakers in the SEA region.}
    \label{fig:sea_map}
    \vspace{-8mm}
\end{wrapfigure}
As shown in Figure~\ref{fig:sea_map}, Southeast Asia is one of the most linguistically and culturally diverse regions in the world, covering languages from multiple language families and writing systems. This diversity creates unique challenges for multilingual multimodal learning, especially because many Southeast Asian languages remain low-resource in both textual and image-text paired data. Recent efforts such as SEACrowd~\cite{seacrowd}, SEA-LION~\cite{sealion}, SEA-HELM~\cite{seahelm}, and SEA-VL~\cite{seavl} have contributed important resources, benchmarks, and models for Southeast Asian language and vision-language research. Nevertheless, these works do not directly address the development of compact and efficient multilingual text-vision embedding models specialized for Southeast Asian languages. Moreover, the low-resource situation in this region is not limited to data; computing resources are also often constrained. Therefore, developing multilingual text-vision embedding models for Southeast Asia requires not only language coverage but also computational efficiency.

In this paper, we study how to build compact multilingual text-vision embedding models for Southeast Asian languages.
We introduce \textbf{SEA-CLIP-Tiny}, a 46.11M-parameter multilingual text-vision embedding model designed to support both English and Southeast Asian languages under limited deployment resources.
Our approach adapts a knowledge distillation framework inspired by CLIP-KD~\cite{yang2024clip}, originally developed for compact CLIP models, to Southeast Asian multilingual settings through multilingual teacher guidance, region-specific training data, and multi-objective distillation.

Our experiments show that standard compact CLIP models trained primarily on general-domain data provide limited retrieval performance for several Southeast Asian languages.
In contrast, \textbf{SEA-CLIP-Tiny} achieves the strongest average retrieval performance among the evaluated student models, reaching 12.9\%, 31.5\%, and 42.2\% at R@1, R@5, and R@10 across seven Southeast Asian languages.
Compared with MobileCLIP2, SEA-CLIP-Tiny improves average R@10 by 12.1 points while using 38.4\% fewer parameters and substantially lower measured CPU latency.
The gains are particularly large for Thai, Myanmar, Malay, and Vietnamese, although MobileCLIP2 remains stronger on several other languages and on CVQA.
Human evaluation further shows that native speakers judge \textbf{SEA-CLIP-Tiny} as competitive with MobileCLIP2, with higher preference rates for SEA-CLIP-Tiny on Thai and Javanese.
%
%
Overall, these results highlight the importance of region-aware adaptation for compact multilingual text-vision models in Southeast Asia.

Our contributions are summarized as follows:
\begin{itemize}

\item We investigate compact multilingual text-vision embedding for Southeast Asia and show that existing compact CLIP-style models trained primarily on general-domain data provide limited support for several SEA languages.

\item We introduce \textbf{SEA-CLIP-Tiny}, a 46.11M-parameter multilingual text--vision embedding model that adapts a CLIP-KD-style framework using multilingual teacher guidance, general-domain and curated Southeast Asian data, and multi-objective distillation.

\item We demonstrate that \textbf{SEA-CLIP-Tiny} achieves the strongest average retrieval performance among the evaluated student models across seven Southeast Asian languages, while using 38.4\% fewer parameters and lower measured CPU latency than MobileCLIP2. Human evaluation and controlled ablations further demonstrate the benefits and limitations of region-specific training.
\end{itemize}
\section{Related Works}
\label{sec:related-works}

\subsection{Multilingual Text-vision Models}

CLIP-style vision-language models align images and texts in a shared embedding space through contrastive learning, enabling zero-shot classification and image-text retrieval~\cite{xiao2025mieb,jiang2024vlm2vec}. However, early models were mostly trained on English image-text pairs, which limits their effectiveness for multilingual and low-resource scenarios. To address this, multilingual CLIP extensions such as mCLIP~\cite{chen-etal-2023-mclip} and AltCLIP~\cite{altclip} align multilingual text encoders with the visual representation space of pretrained CLIP models. These approaches show that multilingual transfer can be achieved without fully retraining large vision-language models from scratch.

Other works explore multilingual knowledge distillation and cross-lingual transfer. For example, mCLIP~\cite{chen-etal-2023-mclip} and NLLB-CLIP~\cite{nllb_clip} use multilingual text encoders, translated captions, or distillation objectives to improve multilingual image-text retrieval under limited data or compute. Language-specific models such as Chinese-CLIP~\cite{chinese_clip} and Italian-CLIP~\cite{bianchi2021contrastivelanguageimagepretrainingitalian} further demonstrate that region- or language-specific image-text data can substantially improve retrieval and zero-shot performance. More recent models, including SigLIP2 and Qwen-VL, expand multilingual multimodal capabilities at larger scales~\cite{siglip2,qwen_vl}. However, these models are typically large or focused on high-resource languages, leaving compact multilingual models for Southeast Asian languages underexplored.

\subsection{Southeast Asian Multimodal Research Landscape}

Recent efforts have significantly advanced NLP and multimodal research for Southeast Asia. SEACrowd collects and standardizes datasets for Southeast Asian languages, while SEA-HELM provides holistic evaluation for large language models in regional linguistic and cultural contexts~\cite{seacrowd,seahelm}. SEA-LION develops open language models for the region, and SEA-VL extends this progress toward vision-language modeling~\cite{sealion,seavl}. In addition, SEA-Guard studies safety for Southeast Asian languages and contexts~\cite{seaguard}.

Together, these works cover data, modeling, evaluation, multimodality, and safety for Southeast Asian AI. However, embedding training remains comparatively less explored. SEA-Embedding~\cite{seaembedding} addresses text-only representation learning and emphasizes the importance of building dedicated embedding models for Southeast Asian languages rather than relying solely on general multilingual models. This insight also applies to text-vision embedding, where both language and visual representations must be aligned. Therefore, we address this gap by proposing an efficient multilingual text-vision embedding model specialized for Southeast Asian languages.

\section{Method}
\label{sec:method}

\subsection{Overview}

\begin{figure}[h]
\centering
\includegraphics[width=0.8\textwidth]{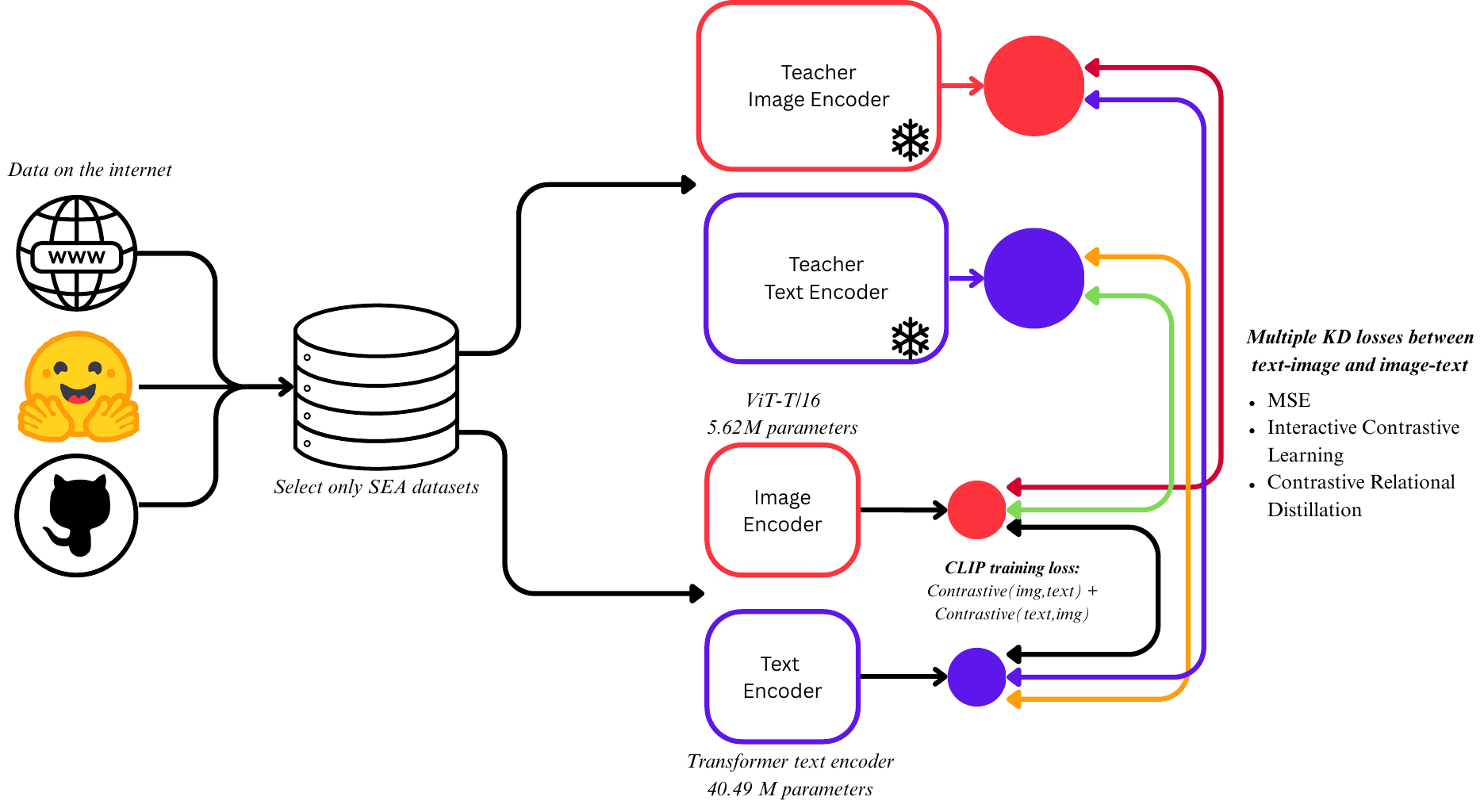}
\vspace{-4mm}
\caption{The overview of \textbf{SEA-CLIP-Tiny}. We split the module into two major components: data selection and training framework.}
\label{fig:overview}
\vspace{-8mm}
\end{figure}

Figure~\ref{fig:overview} summarizes the training methodology of \textbf{SEA-CLIP-Tiny}.
We started with Southeast Asian data selection in Section~\ref{subsec:data_selection}.
Then, we describe how we train \textbf{SEA-CLIP-Tiny} in Section~\ref{subsec:train_model}. 

\subsection{Training Data} \label{subsec:data_selection}

Table~\ref{tab:datasets} summarizes the multilingual training data used for \textbf{SEA-CLIP-Tiny}.
Our SEA-focused data covers all seven languages in the main retrieval evaluation.
We use filtered Cultural Ground Open-Ended~\cite{nyandwi-etal-2025-grounding} (703K pairs), WIT-HF~\cite{srinivasan2021wit} (488K), BLOOM~\cite{leong-etal-2022-bloom} (22K), and SEA-Mammoth~\cite{cahyawijaya2026anthropogenicregionaladaptationmultimodal} (540K).
For SEA-Mammoth, we concatenate the question and answer as the text input and pair it with the corresponding image.
We additionally use CC12M~\cite{changpinyo2021conceptual12mpushingwebscale} as general-domain data, following CLIP-KD~\cite{yang2024clip}.
Our ablations show that SEA-focused data improves multilingual retrieval, while CC12M helps preserve broader visual recognition, motivating their combination in the final training mixture.
The amount of available training data varies substantially across languages, which we discuss as a limitation when interpreting language-level performance.

\begin{table}[h!]
\centering
\vspace{-3mm}
\resizebox{\linewidth}{!}{%
\begin{tabular}{l rrrrrrrr}
\toprule
\textbf{Dataset}
  & \textbf{Indonesian}
  & \textbf{Javanese}
  & \textbf{Malay}
  & \textbf{Sundanese}
  & \textbf{Myanmar}
  & \textbf{Thai}
  & \textbf{Vietnamese}
  & \textbf{Total} \\
\midrule
CG Open-Ended (filtered)~\cite{nyandwi-etal-2025-grounding}
  & 223,098 & 39,747 & 161,397 & 21,238 & ---     & 58,433  & 199,562 & 703,475 \\
WIT-HF~\cite{srinivasan2021wit}
  & 146,744 & 11,926 &  59,504 & ---    &  9,941  & 55,594  & 203,979 & 487,688 \\
BLOOM~\cite{leong-etal-2022-bloom}$^\dagger$
  &   2,169 &    --- &     102 & ---    &    419  &  3,009  &      22 &  21,534 \\
SEA-Mammoth~\cite{cahyawijaya2026anthropogenicregionaladaptationmultimodal}$^\ddagger$
  &   54,040 &    --- &     54,040 & ---    &    54,040  &  54,040  &      54,040 &   540,400 \\
\midrule
\textbf{Total}
  & \textbf{426,051} & \textbf{51,673} & \textbf{275,043} & \textbf{21,238} & \textbf{64,400} & \textbf{171,076} & \textbf{457,603} & \textbf{1,753,097} \\
\bottomrule
\end{tabular}}
\caption{Multilingual training data sources used for \textbf{SEA-CLIP-Tiny}, reported as the number of training examples per target language.
The \textbf{Total} column includes all examples used from each dataset, including additional languages not shown in the table.
$^\dagger$BLOOM covers 76 languages in total.
$^\ddagger$SEA-Mammoth covers 10 languages in total; zh, ta, lo, km, and fil are included only in the Total column.}
\label{tab:datasets}
\vspace{-8mm}
\end{table}

\subsection{Training Framework} \label{subsec:train_model}

\subsubsection{Tokenizer Settings} \label{sec:tokenizer}

Existing compact CLIP models commonly use tokenizers or the teacher model developed primarily from English-centric data~\cite{wu2023tinyclip,yang2024clip}, which can be inefficient for Southeast Asian languages with diverse scripts and segmentation patterns.
We use MetaCLIP-2~\cite{chuang2025metaclip2} with the XLM-V-Base tokenizer as the multilingual teacher, while retaining the CLIP-BPE tokenizer for the compact student to remain compatible with CLIP-KD.
Although the student tokenizer can encode the evaluated SEA languages through byte-level decomposition, its vocabulary contains substantially fewer dedicated SEA-script entries than the teacher tokenizer.
As shown in Table~\ref{tab:tok-vocab}, the student vocabulary contains only 51 Thai-script entries and no dedicated Myanmar, Khmer, or Lao entries, compared with 23,598 SEA-script entries in the XLM-V-Base vocabulary.
This difference leads to substantially different tokenization efficiency across languages.
For Indonesian, Javanese, Malay, and Sundanese, the student produces relatively moderate sequence lengths with truncation below 1\%.
However, Vietnamese, Thai, and Myanmar rely heavily on byte-fragment tokens, resulting in average truncation rates of 38.9\%, 24.8\%, and 95.2\%, respectively, compared with at most 0.8\% for the teacher tokenizer.
Vietnamese, Thai, and Myanmar also show high byte-fragment rates of 50.7\%, 55.9\%, and 99.8\% of emitted student tokens, while the teacher tokenizer emits almost no unknown tokens for any evaluated language.

Across the seven evaluated SEA languages, the student requires 60.4 tokens per caption on average, compared with 20.0 for the teacher (Table~\ref{tab:tokenizer_stats}).
These results show that CLIP-BPE enables multilingual SEA training but remains considerably less token-efficient for non-Latin and heavily diacritized scripts, which we identify as a limitation of the current student model.

\subsubsection{Training Loss} \label{sec:loss}

To train SEA-CLIP-Tiny, we adopt a multi-objective training strategy inspired by CLIP knowledge distillations~\cite{yang2024clip,wu2023tinyclip}.
Our contribution does not introduce a new distillation objective; instead, we apply these objectives with a multilingual teacher and our Southeast Asian training mixture, while previous works~\cite{yang2024clip,wu2023tinyclip} only distilling from a monolingual teacher and data.
Let $\mathcal{B}=\{(I_k,T_k)\}_{k=1}^{|\mathcal{B}|}$ denote a mini-batch of image-text pairs, where $I_k$ is an image and $T_k$ is its paired text, where the texts can be written in multiple languages.
We use a frozen multilingual CLIP model as the teacher, denoted by visual encoder $f^{T}_{v}$ and text encoder $f^{T}_{t}$.
The student model consists of a ViT-T visual encoder $f^{S}_{v}$ and a small multilingual text encoder $f^{S}_{t}$.
For each pair $(I_k,T_k)$, the teacher and student produce visual and textual embeddings, where $\mathrm{norm}(\cdot)$ denotes $\ell_2$ normalization:
\begin{align}
v^{T}_{k} &= \mathrm{norm}(f^{T}_{v}(I_k)), &
s^{T}_{k} &= \mathrm{norm}(f^{T}_{t}(T_k)), \\
v^{S}_{k} &= \mathrm{norm}(f^{S}_{v}(I_k)), &
s^{S}_{k} &= \mathrm{norm}(f^{S}_{t}(T_k)).
\end{align}
When the teacher and student embedding dimensions differ, we apply a learnable linear projection to the student embeddings before normalization so that all embeddings lie in the same $d$-dimensional space.
\noindent
\emph{Mean Squared Error Distillation.}
First, we directly align the student embeddings with the teacher embeddings.
This encourages the student's visual and text encoders to mimic the representation space of the teacher.
The MSE distillation loss is defined as:
\begin{equation}
    \mathcal{L}_{\mathrm{MSE}}
    =
    \frac{1}{|\mathcal{B}|}
    \sum_{k=1}^{|\mathcal{B}|}
    \left(
    \left\| v^{S}_{k} - v^{T}_{k} \right\|^{2}_{2}
    +
    \left\| s^{S}_{k} - s^{T}_{k} \right\|^{2}_{2}
    \right).
\end{equation}

\noindent
\emph{Interactive Contrastive Learning.}
While MSE directly matches the teacher and student embeddings, it does not explicitly use the contrastive structure of image-text pairs.
Therefore, we also use Interactive Contrastive Learning (ICL), where the student embedding is used as the anchor and the teacher embeddings are used as contrastive targets.
For image-to-text alignment, the student image embedding $v^{S}_{k}$ is contrasted against all teacher text embeddings $\{s^{T}_{j}\}_{j=1}^{|\mathcal{B}|}$:
\begin{equation}
    \mathcal{L}^{I \rightarrow T}_{\mathrm{ICL}}
    =
    - \frac{1}{|\mathcal{B}|}
    \sum_{k=1}^{|\mathcal{B}|}
    \log
    \frac{
    \exp\left(v^{S}_{k} \cdot s^{T}_{k} / \tau \right)
    }{
    \sum_{j=1}^{|\mathcal{B}|}
    \exp\left(v^{S}_{k} \cdot s^{T}_{j} / \tau \right)
    },
\end{equation}
where $\tau$ is the temperature parameter.
Similarly, for text-to-image alignment, the student text embedding $s^{S}_{k}$ is contrasted against all teacher image embeddings $\{v^{T}_{j}\}_{j=1}^{|\mathcal{B}|}$:
\begin{equation}
    \mathcal{L}^{T \rightarrow I}_{\mathrm{ICL}}
    =
    - \frac{1}{|\mathcal{B}|}
    \sum_{k=1}^{|\mathcal{B}|}
    \log
    \frac{
    \exp\left(s^{S}_{k} \cdot v^{T}_{k} / \tau \right)
    }{
    \sum_{j=1}^{|\mathcal{B}|}
    \exp\left(s^{S}_{k} \cdot v^{T}_{j} / \tau \right)
    }.
\end{equation}
The final ICL loss is the symmetric average:
\begin{equation}
    \mathcal{L}_{\mathrm{ICL}}
    =
    \frac{1}{2}
    \left(
    \mathcal{L}^{I \rightarrow T}_{\mathrm{ICL}}
    +
    \mathcal{L}^{T \rightarrow I}_{\mathrm{ICL}}
    \right).
\end{equation}

\noindent
\emph{Contrastive Relational Distillation.}
In addition to aligning individual embeddings, we further distill the contrastive relation structure from the teacher to the student. For each image anchor, we first compute the teacher and student image-to-text probability distributions over the mini-batch:
\begin{equation}
p^{T}_{k}[j] = \frac{\exp(v^{T}_{k}\cdot s^{T}_{j}/\tau)}{\sum_{b=1}^{|\mathcal{B}|}\exp(v^{T}_{k}\cdot s^{T}_{b}/\tau)},
\qquad
p^{S}_{k}[j] = \frac{\exp(v^{S}_{k}\cdot s^{S}_{j}/\tau)}{\sum_{b=1}^{|\mathcal{B}|}\exp(v^{S}_{k}\cdot s^{S}_{b}/\tau)}.
\end{equation}
We then minimize the KL divergence between the teacher and student distributions:
\begin{equation}
\mathcal{L}^{I\rightarrow T}_{\mathrm{CRD}} = \frac{1}{|\mathcal{B}|}\sum_{k=1}^{|\mathcal{B}|}\sum_{j=1}^{|\mathcal{B}|} p^{T}_{k}[j]\,\log\frac{p^{T}_{k}[j]}{p^{S}_{k}[j]}.
\end{equation}
The text-to-image term $\mathcal{L}^{T\rightarrow I}_{\mathrm{CRD}}$ is defined analogously using text anchors and the corresponding distributions $q^{T}_{k}$ and $q^{S}_{k}$, and the final CRD loss is $\mathcal{L}_{\mathrm{CRD}} = \mathcal{L}^{I\rightarrow T}_{\mathrm{CRD}} + \mathcal{L}^{T\rightarrow I}_{\mathrm{CRD}}$.

\noindent
\emph{CLIP Training Loss.}
Besides distillation, we train the student model with the standard CLIP contrastive objective so that the student learns image-text alignment from the training data.
Given the student image and text embeddings, the image-to-text contrastive loss is:
\begin{equation}
    \mathcal{L}^{I \rightarrow T}_{\mathrm{CLIP}}
    =
    - \frac{1}{|\mathcal{B}|}
    \sum_{k=1}^{|\mathcal{B}|}
    \log
    \frac{
    \exp\left(v^{S}_{k} \cdot s^{S}_{k} / \tau \right)
    }{
    \sum_{j=1}^{|\mathcal{B}|}
    \exp\left(v^{S}_{k} \cdot s^{S}_{j} / \tau \right)
    }.
\end{equation}
The text-to-image contrastive loss is:
\begin{equation}
    \mathcal{L}^{T \rightarrow I}_{\mathrm{CLIP}}
    =
    - \frac{1}{|\mathcal{B}|}
    \sum_{k=1}^{|\mathcal{B}|}
    \log
    \frac{
    \exp\left(s^{S}_{k} \cdot v^{S}_{k} / \tau \right)
    }{
    \sum_{j=1}^{|\mathcal{B}|}
    \exp\left(s^{S}_{k} \cdot v^{S}_{j} / \tau \right)
    }.
\end{equation}
The symmetric CLIP loss is:
\begin{equation}
    \mathcal{L}_{\mathrm{CLIP}}
    =
    \frac{1}{2}
    \left(
    \mathcal{L}^{I \rightarrow T}_{\mathrm{CLIP}}
    +
    \mathcal{L}^{T \rightarrow I}_{\mathrm{CLIP}}
    \right).
\end{equation}

\noindent
\emph{Overall Training Objective.}
%
The final objective combines the CLIP loss with the three distillation losses, $\mathcal{L} = \mathcal{L}_{\mathrm{CLIP}} + \alpha_{1}\mathcal{L}_{\mathrm{MSE}} + \alpha_{2}\mathcal{L}_{\mathrm{ICL}} + \alpha_{3}\mathcal{L}_{\mathrm{CRD}}$, where $\alpha_{1}, \alpha_{2}, \alpha_{3}$ are loss weights. The teacher encoders are frozen; gradients update only the student encoders and projection layers.
%
%

\section{Experimental Setup}
\label{sec:results}

\subsection{Training Details}

We train \textbf{SEA-CLIP-Tiny} following the hyperparameter settings of CLIP-KD~\cite{yang2024clip}.
We use AdamW with an initial learning rate of $2\times10^{-3}$, weight decay of 0.1, and a linear warm-up over the first 2,000 iterations.
Training is conducted on 40 NVIDIA A100 GPUs with a per-GPU batch size of 128 and 32 epochs.
We use MetaCLIP-2~\cite{chuang2025metaclip2} as the frozen multilingual teacher.
For the student architecture, we use ViT-T/16 and a 12-layer transformer text encoder similar to CLIP-KD, while the student visual encoder, text encoder, and projection layers are updated.
For the contrastive objectives, the temperature $\tau$ is set to 0.07.
The loss weights are $\alpha_{1}=2000$ for MSE, $\alpha_{2}=1$ for ICL, and $\alpha_{3}=1$ for CRD.

\subsection{Benchmarks}

We evaluate \textbf{SEA-CLIP-Tiny} across image-text retrieval, zero-shot classification, and localized visual question answering.
For multilingual retrieval, we use XM3600~\cite{thapliyal-etal-2022-crossmodal}, Flickr30k-200~\cite{nllb_clip}, and XTD-200~\cite{nllb_clip}, which collectively cover seven target SEA languages: Thai, Myanmar, Malay, Indonesian, Javanese, Sundanese, and Vietnamese.
We report the mean of image-to-text and text-to-image Recall@1, @5, and @10.
For language-level evaluation, scores are averaged over the retrieval benchmarks available for each language, while task-level results are reported separately for each benchmark.

Beyond retrieval, we evaluate cultural visual recognition using Babel-IN~\cite{geigle-etal-2024-babel}, reported with zero-shot classification accuracy at @1, @5, and @10.
We also evaluate localized visual question answering using the LOCAL split of CVQA~\cite{mogrovejo2024cvqa}, reported with top-1 accuracy.
For CLIP-style evaluation on CVQA, we rank each candidate by the similarity between the image and the concatenated question-choice text~\footnote{We perform a similarity comparison between \{Image\} \& CONCAT\{\{Question\} + \{Choice\}\} following previous vision-text embedding works~\cite{mogrovejo2024cvqa,cahyawijaya2026anthropogenicregionaladaptationmultimodal}.} using CVQA~\cite{mogrovejo2024cvqa} (top-1 accuracy on the LOCAL split).
Because retrieval, classification, and VQA use different metrics and language coverage, we report them separately rather than combining them into a single aggregate score.
We acknowledge that SEA benchmark availability remains limited, and some existing benchmarks rely on translated data.
While machine-translated evaluation can introduce artifacts for low-resource SEA languages~\cite{seacrowd,seahelm,seaembedding}, these datasets remain useful for evaluating languages with limited native benchmark resources.
We therefore report results across multiple established benchmarks and discuss this limitation when interpreting the results.
Please see Table~\ref{tab:coverage} for the full benchmark coverage details.

\subsection{Competitive methods} 

We compare \textbf{SEA-CLIP-Tiny} against three compact student baselines and the multilingual teacher.
We report MetaCLIP-2 (ViT-B-16-worldwide@WorldWideCLIP)~\cite{chuang2025metaclip2} as the teacher reference to measure the remaining teacher-student performance gap.
For a controlled compact baseline, we evaluate CLIP-KD~\cite{yang2024clip} trained on CC12M, which uses the same student architecture as \textbf{SEA-CLIP-Tiny} but without our Southeast Asian training mixture.
We additionally compare against publicly released TinyCLIP~\cite{wu2023tinyclip} and MobileCLIP2~\cite{vasu2024mobileclip}, representing existing efficient CLIP variants.
TinyCLIP is trained on CC12M, while MobileCLIP2 uses the substantially larger DFN-2B dataset~\cite{fang2024data}.
These baselines, therefore, differ in both training data and training recipe, and we use them as comparisons with existing pretrained compact models rather than as fully controlled experiments.
For \textbf{SEA-CLIP-Tiny}, we report the final model trained with CC12M and the full Southeast Asian data mixture described in Section~\ref{subsec:data_selection}.

\section{Experimental Results}

\subsection{Main Results: Language Score}
\textbf{Goal.}
Our goal is to evaluate the multilingual retrieval ability of \textbf{SEA-CLIP-Tiny} across Southeast Asian languages. 
We compare our model with existing compact CLIP-style baselines, including CLIP-KD, TinyCLIP, and MobileCLIP2, as well as a strong multilingual teacher model. 
We evaluate seven Southeast Asian languages: Thai, Myanmar, Malay, Indonesian, Javanese, Sundanese, and Vietnamese. 
Performance is reported at R@1, R@5, and R@10 using mean image-to-text
and text-to-image retrieval.
For each language, we average over the retrieval benchmarks that cover that language: XM3600, Flickr30k-200, and XTD-200 for Thai, Indonesian, and Vietnamese, and Flickr30k-200 and XTD-200 for the remaining languages.
We also report the number of image and text encoder parameters and CPU ms to compare retrieval performance under different model sizes.

\begin{table}[h!]
\vspace{-3mm}
\centering\setlength{\tabcolsep}{3pt}
\resizebox{\linewidth}{!}{%
\begin{tabular}{l c c ccc ccc ccc ccc ccc ccc ccc ccc}
\toprule
\multirow{2}{*}{\textbf{Model}} & \multirow{2}{*}{\textbf{\#Params (M)}} & \multirow{2}{*}{\textbf{CPU ms}}
  & \multicolumn{3}{c}{\textbf{Thai}}
  & \multicolumn{3}{c}{\textbf{Myanmar}}
  & \multicolumn{3}{c}{\textbf{Malay}}
  & \multicolumn{3}{c}{\textbf{Indonesian}}
  & \multicolumn{3}{c}{\textbf{Javanese}}
  & \multicolumn{3}{c}{\textbf{Sundanese}}
  & \multicolumn{3}{c}{\textbf{Vietnamese}}
  & \multicolumn{3}{c}{\textbf{Avg}} \\
\cmidrule(lr){4-6}\cmidrule(lr){7-9}\cmidrule(lr){10-12}\cmidrule(lr){13-15}\cmidrule(lr){16-18}\cmidrule(lr){19-21}\cmidrule(lr){22-24}\cmidrule(lr){25-27}
  & & & @1 & @5 & @10 & @1 & @5 & @10 & @1 & @5 & @10 & @1 & @5 & @10 & @1 & @5 & @10 & @1 & @5 & @10 & @1 & @5 & @10 & @1 & @5 & @10 \\
\midrule
Teacher & $86.19+499.77$ & \textit{90.5}
  & \textit{48.7} & \textit{75.5} & \textit{84.7}
  & \textit{25.3} & \textit{49.1} & \textit{60.6}
  & \textit{56.1} & \textit{81.9} & \textit{89.5}
  & \textit{59.9} & \textit{84.2} & \textit{91.2}
  & \textit{27.4} & \textit{51.8} & \textit{63.6}
  & \textit{24.5} & \textit{47.6} & \textit{58.8}
  & \textit{57.1} & \textit{83.5} & \textit{90.7}
  & \textit{42.7} & \textit{67.6} & \textit{77.0} \\
\midrule
CLIP-KD & $5.62+40.49$ & 24.1
  & 0.2 & 0.6 & 1.0
  & 0.4 & 1.3 & 1.9
  & 1.9 & 5.2 & 7.5
  & 2.6 & 6.9 & 9.7
  & 2.9 & 8.6 & 12.6
  & 4.0 & 9.9 & 15.2
  & 0.6 & 1.6 & 2.4
  & 1.8 & 4.9 & 7.2 \\
TinyCLIP & $8.28+15.17$ & 14.9
  & 0.1 & 0.5 & 1.0
  & 0.7 & 1.9 & 2.5
  & 2.5 & 6.1 & 8.9
  & 3.0 & 8.2 & 11.4
  & 3.7 & 9.3 & 13.4
  & 4.4 & 11.2 & 16.5
  & 1.3 & 3.6 & 5.2
  & 2.2 & 5.8 & 8.4 \\
MobileCLIP2 & $11.41+63.43$ & 61.6
  & 0.9 & 1.9 & 2.7
  & 1.7 & 3.8 & 4.8
  & 20.9 & 40.6 & 50.8
  & \textbf{30.1} & \textbf{53.4} & \textbf{64.1}
  & \textbf{13.6} & \textbf{30.4} & \textbf{39.8}
  & \textbf{14.0} & \textbf{29.7} & \textbf{39.5}
  & 3.3 & 7.2 & 9.3
  & 12.1 & 23.9 & 30.1 \\
\midrule
\textbf{SEA-CLIP-Tiny} & $5.62+40.49$ & 24.1
  & \textbf{8.6} & \textbf{24.1} & \textbf{33.9}
  & \textbf{3.6} & \textbf{12.1} & \textbf{18.7}
  & \textbf{22.6} & \textbf{49.3} & \textbf{63.2}
  & 22.5 & 49.0 & 62.5
  & 9.8 & 25.4 & 35.8
  & 7.1 & 20.8 & 29.2
  & \textbf{16.4} & \textbf{39.6} & \textbf{52.4}
  & \textbf{12.9} & \textbf{31.5} & \textbf{42.2} \\
\bottomrule\end{tabular}
}
\caption{Per-language image-text retrieval performance, reported as \%. For each language, we report mean(I2T, T2I) R@$k$, averaged over the retrieval benchmarks covering that language. \textbf{CPU ms} denotes median single-sample CPU latency for encoding one image and one caption; full computational measurements are reported in Tables~\ref{tab:efficiency} and~\ref{tab:efficiency-cpu}. \textbf{Bold} denotes the best student model in each column.}
\vspace{-8mm}
\label{tab:per_language}
\end{table}

\noindent
\textbf{Discussion.}
Table~\ref{tab:per_language} reports per-language retrieval performance across seven SEA languages at R@1, R@5, and R@10. \textbf{SEA-CLIP-Tiny} achieves the best average performance among the evaluated student models at all retrieval ranks, reaching 12.9\%, 31.5\%, and 42.2\% at R@1, R@5, and R@10, respectively.
Compared with MobileCLIP2, the strongest compact baseline, this corresponds to gains of 0.8, 7.6, and 12.1 points, respectively.
SEA-CLIP-Tiny also uses 38.4\% fewer parameters and reduces the measured CPU latency from 61.6 ms to 24.1 ms.
The gains are particularly large for Thai, Myanmar, and Vietnamese, where SEA-CLIP-Tiny outperforms MobileCLIP2 at all retrieval ranks.
At R@10, performance improves from 2.7\% to 33.9\% for Thai, from 4.8\% to 18.7\% for Myanmar, and from 9.3\% to 52.4\% for Vietnamese.
SEA-CLIP-Tiny also improves Malay R@5 from 40.6\% to 49.3\% and R@10 from 50.8\% to 63.2\%.
In contrast, MobileCLIP2 remains stronger on Indonesian, Javanese, and Sundanese.

These results suggest that SEA-focused training substantially improves retrieval coverage for languages that are poorly supported by existing compact models.
At the same time, the language-level differences show that the gains are not uniform across Southeast Asia.
We therefore characterize SEA-CLIP-Tiny as the strongest average-performing student among the evaluated models, rather than claiming consistent superiority for every language.
Additional I2T and T2I results are reported in Supplementary~\ref{appendix:it2_t2i}.

\subsection{Main Results: Task Score}

\noindent
\textbf{Goal.}
Our goal is to evaluate whether \textbf{SEA-CLIP-Tiny} improves performance across different Southeast Asian vision-language tasks while remaining compact.
We compare against compact CLIP-style baselines and a strong multilingual teacher on three image-text retrieval benchmarks (XM3600, Flickr30k-200, and XTD-200), zero-shot classification on Babel-IN, and localized visual question answering on CVQA.
We keep the metrics for retrieval, classification, and VQA separate to avoid conflating different task families.

\begin{table}[h!]\centering\setlength{\tabcolsep}{3pt}
\vspace{-3mm}
\resizebox{\linewidth}{!}{%
\begin{tabular}{l c c ccc ccc ccc ccc ccc c}
\toprule
\multirow{3}{*}{\textbf{Model}} & \multirow{3}{*}{\textbf{\#Params (M)}} & \multirow{3}{*}{\textbf{CPU ms}} & \multicolumn{12}{c}{\textbf{Retrieval} (R@$k$)} & \multicolumn{3}{c}{\textbf{Classification} (acc@$k$)} & \textbf{VQA} \\
\cmidrule(lr){4-15}\cmidrule(lr){16-18}\cmidrule(lr){19-19}
  & & & \multicolumn{3}{c}{XM3600} & \multicolumn{3}{c}{Flickr30k-200} & \multicolumn{3}{c}{XTD-200} & \multicolumn{3}{c}{\textbf{Avg}} & \multicolumn{3}{c}{Babel-IN} & CVQA \\
\cmidrule(lr){4-6}\cmidrule(lr){7-9}\cmidrule(lr){10-12}\cmidrule(lr){13-15}\cmidrule(lr){16-18}
  & & & @1 & @5 & @10 & @1 & @5 & @10 & @1 & @5 & @10 & @1 & @5 & @10 & @1 & @5 & @10 & acc@1 \\
\midrule
Teacher & $86.19+499.77$ & \textit{90.5}
  & \textit{55.7} & \textit{81.4} & \textit{88.8} & \textit{46.0} & \textit{69.3} & \textit{77.0} & \textit{39.2} & \textit{65.8} & \textit{77.0} & \textit{47.0} & \textit{72.2} & \textit{81.0} & \textit{42.3} & \textit{61.3} & \textit{67.8} & \textit{48.6} \\
\midrule
CLIP-KD & $5.62+40.49$ & 24.1
  & 1.1 & 2.6 & 3.5 & 1.5 & 3.7 & 5.4 & 2.2 & 6.2 & 9.4 & 1.6 & 4.2 & 6.1 & 4.2 & 7.7 & 10.5 & 25.9 \\
TinyCLIP & $8.28+15.17$ & 14.9
  & 1.6 & 3.7 & 4.9 & 2.1 & 4.7 & 6.8 & 2.4 & 7.1 & 10.5 & 2.0 & 5.2 & 7.4 & 4.4 & 8.2 & 10.8 & 26.8 \\
MobileCLIP2 & $11.41+63.43$ & 61.6
  & 11.7 & 20.8 & 24.6 & 11.2 & 22.0 & 27.5 & 12.9 & 25.8 & 33.1 & 11.9 & 22.8 & 28.4 & 15.6 & 23.5 & 27.5 & \textbf{33.5} \\
\midrule
SEA-CLIP-Tiny & $5.62+40.49$ & 24.1
  & \textbf{16.0} & \textbf{37.4} & \textbf{48.6} & \textbf{11.3} & \textbf{26.6} & \textbf{35.1} & \textbf{14.4} & \textbf{36.4} & \textbf{49.8} & \textbf{13.9} & \textbf{33.5} & \textbf{44.5} & \textbf{17.2} & \textbf{33.3} & \textbf{41.1} & 28.2 \\
\bottomrule\end{tabular}
}
\caption{Per-task SEA performance, reported as \%. Retrieval reports mean(I2T, T2I) R@$k$ on XM3600, Flickr30k-200, and XTD-200, with \textbf{Avg} computed only across these three retrieval benchmarks.
Classification reports Babel-IN zero-shot acc@$k$, and VQA reports CVQA acc@1 on the LOCAL split. \textbf{Bold} denotes the best student model in each column.}
\vspace{-8mm}
\label{tab:per_task}
\end{table}

\noindent
\textbf{Discussion.}
Table~\ref{tab:per_task} shows that \textbf{SEA-CLIP-Tiny} achieves the strongest average retrieval performance among the evaluated student models.
Compared with MobileCLIP2, the strongest compact baseline, the retrieval average improves from 11.9\% to 13.9\% at R@1, from 22.8\% to 33.5\% at R@5, and from 28.4\% to 44.5\% at R@10.
The largest gains appear at higher retrieval ranks, with improvements of 10.7 and 16.1 points at R@5 and R@10, respectively.
SEA-CLIP-Tiny also outperforms MobileCLIP2 on each individual retrieval benchmark at all three ranks.
The improvement extends to zero-shot classification on Babel-IN, where SEA-CLIP-Tiny reaches 17.2\%, 33.3\%, and 41.1\% at acc@1, acc@5, and acc@10, respectively, compared with 15.6\%, 23.5\%, and 27.5\% for MobileCLIP2.
In contrast, MobileCLIP2 remains stronger on CVQA, achieving 33.5\% compared with 28.2\%.
This difference suggests that improved image-text alignment does not necessarily transfer equally to localized VQA.
SEA-CLIP-Tiny achieves these results with 46.11M parameters and 24.1 ms measured CPU latency, compared with 74.84M parameters and 61.6 ms for MobileCLIP2.
Overall, the results show that SEA-focused training substantially improves retrieval and classification for a compact model, while a clear gap remains on more complex tasks such as VQA.

\subsection{Human Preference Scores}

\noindent
\textbf{Setup.}
We conduct a human evaluation to compare the retrieval quality of \textbf{SEA-CLIP-Tiny} against MobileCLIP2 from the perspective of native speakers. This study is motivated by the per-language automatic results in Table~\ref{tab:per_language}, where MobileCLIP2 performs better than our model on several languages, while our model performs better on others. Therefore, we select Indonesian, Javanese, and Sundanese, where MobileCLIP2 achieves stronger automatic scores, and Thai, where \textbf{SEA-CLIP-Tiny} clearly outperforms MobileCLIP2.
For each language, we sample 50 examples from Flickr30k-200 and 50 examples from XTD-200, resulting in 100 evaluation samples per language. For each image query, annotators are shown two candidate retrieval outputs, one produced by \textbf{SEA-CLIP-Tiny} and the other by MobileCLIP2, anonymized as Model A and Model B with randomized order to reduce positional bias. Annotators are asked to select which output is more correct and natural for a native speaker of the target language, or mark them as tied if both are of similar quality. The results of each dataset separately and the annotator guidelines and details are discussed in Supplementary~\ref{appendix:human}.

\begin{figure}[h!]
    \centering
    \vspace{-8mm}
    \includegraphics[width=0.6\linewidth]{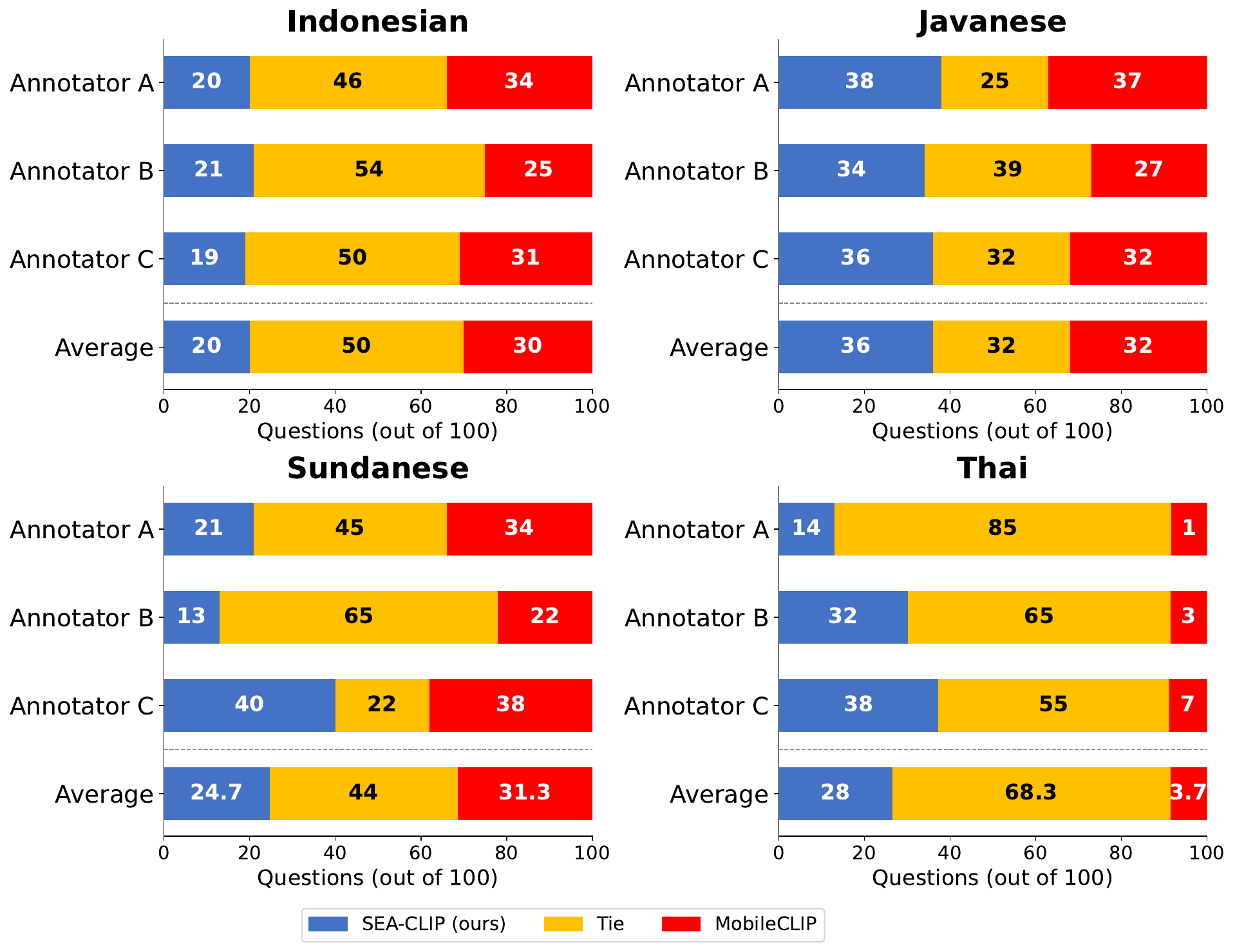}
    \vspace{-4mm}
    \caption{Human preference evaluation comparing SEA-CLIP-Tiny against MobileCLIP2 across four Southeast Asian languages (Indonesian, Javanese, Thai, Sundanese). Each bar shows the number of questions (out of 100) where SEA-CLIP wins (blue), ties (yellow), or MobileCLIP wins (red), per annotator and on average.}
    \vspace{-6mm}
    \label{fig:he_total}
\end{figure}

\noindent
\textbf{Discussion.}
Figure~\ref{fig:he_total} shows that human preference does not always reflect the same gap observed in automatic retrieval scores. Although Table~\ref{tab:per_language} shows that MobileCLIP2 achieves stronger automatic scores on Indonesian, Javanese, and Sundanese, native-speaker judgments suggest that the perceived quality difference is smaller. In Javanese, \textbf{SEA-CLIP-Tiny} slightly outperforms MobileCLIP2, winning 36\% of examples compared to 32.7\% for MobileCLIP2, while 32.7\% are marked as ties. In Sundanese, MobileCLIP2 wins more often, but the gap is moderate, with 24.7\% wins for \textbf{SEA-CLIP-Tiny}, 31.3\% wins for MobileCLIP2, and 44\% ties. Indonesian shows a larger preference for MobileCLIP2, but ties still account for 50\% of the comparisons.
In contrast, for Thai, where MobileCLIP2 performs poorly in the automatic evaluation, \textbf{SEA-CLIP-Tiny} is clearly preferred by human annotators. Our model wins 28\% of the comparisons, while MobileCLIP2 wins only 3.7\%, with the remaining 68.3\% marked as ties. 
Overall, these results indicate that \textbf{SEA-CLIP-Tiny} is competitive with MobileCLIP2 in languages where automatic metrics favor MobileCLIP2, and can outperform it in human preference on Javanese and Thai.

\section{Ablation Studies} \label{sec:analysis}

\noindent
\textbf{Setup.}
We conduct ablation studies to analyze the contribution of training data and loss design in \textbf{SEA-CLIP-Tiny}. 
All models use the same architecture, training pipeline, and hyperparameters, while varying either the data mixture or the training objective. 
For dataset ablation, we compare the full model, which uses CC12M, curated SEA datasets, and Mammoth, against variants trained with only SEA data, only CC12M, CC12M+SEA without Mammoth, and single-source SEA datasets. 
For loss ablation, we use the same training with the best results from the ablation study and compare the full objective against variants that remove KD losses or use only one loss component. 
We report source-specific retrieval performance, ImageNet accuracy, and the retrieval R@1 average over XM3600, Flickr30k-200, and XTD-200.

\begin{table}[h!]
\centering
\vspace{-3mm}
\resizebox{\linewidth}{!}{%
\begin{tabular}{l l c c c c c}
\toprule
\textbf{Setup} & \textbf{Train data (\#sample)}
  & \textbf{CG R@1} & \textbf{WIT R@1} & \textbf{Bloom R@1}
  & \textbf{ImageNet} & \textbf{R@1-Avg} \\
\midrule
SEA-CLIP-Tiny
  & CC12M + CG-OE-filt + WIT-HF + Bloom + Mammoth \textit{(12.72M)}
 & 29.3 & 21.8 & 23.3 & 36.4 & \textbf{13.9}  \\ \midrule
\multicolumn{7}{c}{\textit{Dataset Ablation}} \\ \midrule
Only SEA datasets
  & CG-OE-filt + WIT-HF + Bloom \textit{(1.21M)}
  & \textbf{63.7} & \textbf{54.6} & \textbf{24.9} & 5.6 & {5.9} \\
Only CC12M
  & CC12M (10.97M)
  & 1.0 & 1.9 & 1.2 & \textbf{40.0} & 6.5 \\
Only WIT
  & WIT-hf (487K)
  & 2.0 & 27.7 & 2.2 & 1.3 & 6.2 \\
Only Bloom
  & Bloom-only (21K)
  & 0.0 & 0.0 & 1.0 & 0.1 & 5.2 \\
Only CG
  & CG-only (703K)
  & 46.5 & 0.2 & 0.6 & 0.2 & 5.5 \\
Only Mammoth
  & SEA-Mammoth (540K)
  & 0.0 & 0.0 & 0.8 & 0.5 & 5.2  \\
CC12M + SEA
  & CC12M + CG-OE-filt + WIT-HF + Bloom \textit{(12.18M)}
  & 33.3 & 25.2 & 16.1 & 37.4 & {9.1} \\
\midrule
\multicolumn{7}{c}{\textit{KD Loss Ablation}} \\ \midrule
Only CLIP 
  & CC12M + CG-OE-filt + WIT-HF + Bloom + Mammoth \textit{(12.72M)}
  & 39.1 & 26.1 & {15.6} & 31.4 & 7.3 \\
CLIP + MSE
  & CC12M + CG-OE-filt + WIT-HF + Bloom + Mammoth \textit{(12.72M)}
  & {39.5} & {26.8} & 11.3 & 31.3 & 7.2 \\
CLIP + ICL
  & CC12M + CG-OE-filt + WIT-HF + Bloom + Mammoth \textit{(12.72M)}
  & 39.2 & 26.4 & 15.3 & {34.2} & {12.2} \\
CLIP + CRD
  & CC12M + CG-OE-filt + WIT-HF + Bloom + Mammoth \textit{(12.72M)}
  & 28.5 & 20.3 & 11.3 & 32.7 & 9.3 \\
\bottomrule
\end{tabular}%
}
\caption{Controlled ablations of training data and distillation objectives for
\textbf{SEA-CLIP-Tiny}, reported as \%.
All variants use the same student architecture and hyperparameters.
\textbf{Retrieval Avg R@1} is the mean retrieval R@1 over XM3600,
Flickr30k-200, and XTD-200, consistent with Table~\ref{tab:per_task}. Note that we use the same multilingual teacher model, \texttt{ViT-B-16-worldwide@WorldWideCLIP}.}

\label{tab:training}
\vspace{-8mm}
\end{table}

\noindent
\textbf{Dataset Ablation.}
Table~\ref{tab:training} shows that the full data mixture provides the strongest downstream SEA performance. \textbf{SEA-CLIP-Tiny}, trained with CC12M, curated SEA datasets, and Mammoth, achieves the highest \texttt{R@1-Avg} of 13.9\%, improving over CC12M-only, SEA-only, Mammoth-only, and CC12M+SEA. In contrast, CC12M-only obtains the best ImageNet score at 40.0\%, but performs poorly on the SEA training-source probes and reaches only 6.5\% \texttt{R@1-Avg}. SEA-only training gives strong source-specific alignment, especially on CG and WIT, but its ImageNet and \texttt{R@1-Avg} scores remain low. These results suggest that general data helps preserve broad visual recognition, while curated SEA data and Mammoth improve regional multilingual retrieval.
The single-source ablations further show that no individual SEA dataset is sufficient by itself. WIT-only improves WIT alignment, and CG-only improves CG alignment, but both transfer weakly to the overall SEA evaluation. Overall, the dataset ablation confirms that strong SEA retrieval performance requires a diverse mixture of general and regional data sources.

\noindent
\textbf{KD Loss Ablation.}
We evaluate the contribution of each KD loss component using the same full training data mixture. Removing KD losses (only CLIP) reduces \texttt{R@1-Avg} from 13.9\% to 7.3\%, showing that KD losses are important for improving downstream generalization. Among the single-loss variants, using CLIP + ICL gives the strongest result, reaching 12.2\% \texttt{R@1-Avg}, while CLIP + CRD and CLIP + MSE achieve 9.3\% and 7.2\%, respectively. However, all single-loss variants remain below the full \textbf{SEA-CLIP-Tiny}. This suggests that each loss contributes different supervision signals, and combining multiple KD losses provides the best balance between preserving general visual-language knowledge and improving SEA retrieval performance.

\section{Conclusion}

In this paper, we introduced \textbf{SEA-CLIP-Tiny}, a compact multilingual text-vision embedding model for Southeast Asian languages.
Our model adapts a CLIP-KD-style framework with Southeast Asian data, multilingual teacher guidance, and multi-objective distillation.
Across retrieval, classification, VQA, human evaluation, and ablation studies, SEA-CLIP-Tiny achieves the strongest average retrieval performance among the evaluated student models, with large gains on under-supported languages.
It also uses 38.4\% fewer parameters and lower CPU latency than MobileCLIP2, although MobileCLIP2 remains stronger on CVQA.
Our ablations further show that general-domain and regional data are complementary, highlighting the importance of region-aware training for compact multilingual vision-language models.
%
%

\section*{Limitation, Broader Impact, and Ethical Discussion}

SEA-CLIP-Tiny improves compact multilingual vision-language retrieval for Southeast Asian languages, but still has a gap to large teacher models and does not cover all languages, dialects, and cultural variations in the region. The model may also inherit biases from its training data and teacher models.
This work aims to make multilingual vision-language models more accessible for Southeast Asia through a compact and efficient model. However, it should be used carefully in high-stakes or culturally sensitive applications.
For human evaluation, annotators are volunteer undergraduate students from Southeast Asia who are native or fluent speakers of the evaluated languages. Participation is voluntary, and results are reported only in aggregate form without releasing personally identifiable information.

\section*{Acknowledgment}

This project is supported by the National Research Foundation, Singapore under its National Large Language Models Funding Initiative. Any opinions, findings and conclusions or recommendations expressed in this material are those of the author(s) and do not reflect the views of National Research Foundation, Singapore. 
This work was also supported in part by WangchanX, an open-source initiative of VISTEC funded by PTT, SCB, and SCBX. We are also grateful for the computational resources supported by the NSTDA Supercomputer.

%
%
\bibliographystyle{splncs04}
\bibliography{main}

\newpage
\section{Supplementary}

\subsection{I2T and T2I Results}  \label{appendix:it2_t2i}
\noindent
\textbf{Discussion.}
The additional I2T and T2I results show that the main trend is consistent across retrieval directions. \textbf{SEA-CLIP-Tiny (CC12M+SEA)} substantially improves over both CC12M-only and SEA-only variants, confirming that general and regional data are complementary. Compared with MobileCLIP2, our model is especially competitive at higher retrieval ranks and shows clear gains for under-supported languages such as Thai, Myanmar, and Vietnamese, while MobileCLIP2 remains stronger in some high-resource languages and at R@1. These results further support the effectiveness of targeted SEA data for compact multilingual retrieval.

\begin{table}[h]
\vspace{-6mm}
\centering\setlength{\tabcolsep}{3pt}
\resizebox{\linewidth}{!}{%
\begin{tabular}{l c c ccc ccc ccc ccc ccc ccc ccc ccc}
\toprule
\multirow{2}{*}{\textbf{Model}} & \multirow{2}{*}{\textbf{\#Params (M)}} & \multirow{2}{*}{\textbf{CPU ms}}
  & \multicolumn{3}{c}{\textbf{Thai}}
  & \multicolumn{3}{c}{\textbf{Myanmar$^\dagger$}}
  & \multicolumn{3}{c}{\textbf{Malay$^\dagger$}}
  & \multicolumn{3}{c}{\textbf{Indonesian}}
  & \multicolumn{3}{c}{\textbf{Javanese$^\dagger$}}
  & \multicolumn{3}{c}{\textbf{Sundanese$^\dagger$}}
  & \multicolumn{3}{c}{\textbf{Vietnamese}}
  & \multicolumn{3}{c}{\textbf{Avg}} \\
\cmidrule(lr){4-6}\cmidrule(lr){7-9}\cmidrule(lr){10-12}\cmidrule(lr){13-15}\cmidrule(lr){16-18}\cmidrule(lr){19-21}\cmidrule(lr){22-24}\cmidrule(lr){25-27}
  & & & @1 & @5 & @10 & @1 & @5 & @10 & @1 & @5 & @10 & @1 & @5 & @10 & @1 & @5 & @10 & @1 & @5 & @10 & @1 & @5 & @10 & @1 & @5 & @10 \\
\midrule
Teacher & $86.19+499.77$ & \textit{90.5}
  & \textit{50.7} & \textit{77.2} & \textit{86.4}
  & \textit{27.2} & \textit{51.1} & \textit{62.3}
  & \textit{57.4} & \textit{82.8} & \textit{90.5}
  & \textit{61.3} & \textit{85.7} & \textit{91.9}
  & \textit{28.1} & \textit{52.7} & \textit{63.9}
  & \textit{24.6} & \textit{48.2} & \textit{59.1}
  & \textit{59.2} & \textit{85.5} & \textit{91.9}
  & \textit{44.1} & \textit{69.0} & \textit{78.0} \\
\midrule
CLIP-KD$^{\ast}$ & $5.62+40.49$ & 24.1
  & 0.2 & 0.7 & 1.1
  & 0.8 & 1.7 & 2.2
  & 2.4 & 6.2 & 8.8
  & 3.0 & 8.2 & 11.4
  & 3.6 & 10.0 & 14.4
  & 4.9 & 11.4 & 17.1
  & 1.0 & 2.6 & 3.7
  & 2.3 & 5.8 & 8.4 \\
TinyCLIP & $8.28+15.17$ & 14.9
  & 0.2 & 0.6 & 0.8
  & 1.2 & 2.4 & 3.3
  & 3.4 & 7.8 & 10.8
  & 3.9 & 10.1 & 13.6
  & 4.7 & 11.7 & 15.9
  & 5.5 & 13.3 & 18.6
  & 1.8 & 4.9 & 6.7
  & 3.0 & 7.2 & 10.0 \\
MobileCLIP2 & $11.41+63.43$ & 61.6
  & 1.2 & 2.5 & 3.6
  & 2.7 & 5.2 & 6.1
  & \textbf{24.1} & 46.1 & 56.8
  & \textbf{34.4} & \textbf{60.2} & \textbf{71.2}
  & \textbf{16.7} & \textbf{35.5} & \textbf{45.7}
  & \textbf{15.8} & \textbf{33.6} & \textbf{43.3}
  & 4.3 & 9.2 & 11.9
  & \textbf{14.2} & 27.5 & 34.1 \\
\midrule
\textbf{SEA-CLIP-Tiny} & $5.62+40.49$ & 24.1
  & \textbf{9.1} & \textbf{25.5} & \textbf{35.4}
  & \textbf{3.8} & \textbf{13.2} & \textbf{19.7}
  & 23.7 & \textbf{51.9} & \textbf{66.2}
  & 23.6 & 50.1 & 63.7
  & 10.7 & 26.2 & 36.6
  & 7.5 & 21.6 & 30.2
  & \textbf{17.4} & \textbf{41.6} & \textbf{54.6}
  & 13.7 & \textbf{32.9} & \textbf{43.7} \\
\bottomrule
\end{tabular}
}
\caption{Per-language performance (mean across XM3600, Flickr30k-200, XTD-200 I2T R@k), reported as \%. Bold = best among student models per column. $\dagger$~Averaged over Flickr30k-200 and XTD-200 only; XM3600 does not cover these languages. $*$~Retrained by on CC12M following CLIPKD's scripts.}
\label{tab:per_language_rk_i2t}
\vspace{-8mm}

\end{table}

\begin{table}[h!]\centering\setlength{\tabcolsep}{3pt}
\vspace{-8mm}
\resizebox{\linewidth}{!}{%
\begin{tabular}{l c c ccc ccc ccc ccc ccc c}
\toprule
\multirow{3}{*}{\textbf{Model}} & \multirow{3}{*}{\textbf{\#Params (M)}} & \multirow{3}{*}{\textbf{CPU ms}} & \multicolumn{12}{c}{\textbf{Retrieval} (R@$k$)} & \multicolumn{3}{c}{\textbf{Classification} (acc@$k$)} & \textbf{VQA} \\
\cmidrule(lr){4-15}\cmidrule(lr){16-18}\cmidrule(lr){19-19}
  & & & \multicolumn{3}{c}{XM3600} & \multicolumn{3}{c}{Flickr30k-200} & \multicolumn{3}{c}{XTD-200} & \multicolumn{3}{c}{\textbf{Avg}} & \multicolumn{3}{c}{Babel-IN} & CVQA \\
\cmidrule(lr){4-6}\cmidrule(lr){7-9}\cmidrule(lr){10-12}\cmidrule(lr){13-15}\cmidrule(lr){16-18}
  & & & @1 & @5 & @10 & @1 & @5 & @10 & @1 & @5 & @10 & @1 & @5 & @10 & @1 & @5 & @10 & acc@1 \\
\midrule
Teacher & $86.19+499.77$ & \textit{90.5}
  & \textit{58.9} & \textit{84.4} & \textit{90.9} & \textit{47.1} & \textit{70.6} & \textit{78.0} & \textit{40.3} & \textit{66.8} & \textit{77.6} & \textit{48.7} & \textit{73.9} & \textit{82.2} & \textit{42.3} & \textit{61.3} & \textit{67.8} & \textit{48.6} \\
\midrule
CLIP-KD$^{\ast}$ & $5.62+40.49$ & 24.1
  & 1.5 & 3.4 & 4.4 & 1.8 & 4.4 & 6.3 & 2.7 & 7.4 & 10.9 & 2.0 & 5.1 & 7.2 & 4.2 & 7.7 & 10.5 & 25.9 \\
TinyCLIP & $8.28+15.17$ & 14.9
  & 2.1 & 4.7 & 5.9 & 2.7 & 5.8 & 8.0 & 3.1 & 8.9 & 12.4 & 2.6 & 6.5 & 8.8 & 4.4 & 8.2 & 10.8 & 26.8 \\
MobileCLIP2 & $11.41+63.43$ & 61.6
  & 14.2 & 24.2 & 28.5 & \textbf{13.2} & 25.9 & 31.8 & 14.8 & 28.9 & 36.6 & 14.0 & 26.3 & 32.3 & 15.6 & 23.5 & 27.5 & \textbf{33.5} \\
\midrule
SEA-CLIP-Tiny & $5.62+40.49$ & 24.1
  & \textbf{16.7} & \textbf{38.6} & \textbf{50.0} & 11.9 & \textbf{28.0} & \textbf{36.4} & \textbf{15.5} & \textbf{37.9} & \textbf{51.6} & \textbf{14.7} & \textbf{34.9} & \textbf{46.0} & \textbf{17.2} & \textbf{33.3} & \textbf{41.1} & 28.2 \\
\bottomrule\end{tabular}
}
\caption{Per-task SEA performance, reported as \%.
  Babel-IN = acc@$k$ (SEA-lang avg); XM3600, Flickr30k-200, XTD-200 = I2T (image$\to$text) R@$k$;
  CVQA = acc@1 (LOCAL, SEA-7 subset).
  \textbf{Retrieval Avg} = mean(Babel-IN, XM3600, Flickr30k-200, XTD-200).
  \textbf{Total Avg} = equal-weight mean over all 5 tasks.
  \textbf{Bold} = best student per column.}
  \vspace{-12mm}
\label{tab:per_task_rk_i2t}
\end{table}

\vspace{-3mm}
\begin{table}[h!]\centering\setlength{\tabcolsep}{3pt}
\resizebox{\linewidth}{!}{%
\begin{tabular}{l c c ccc ccc ccc ccc ccc ccc ccc ccc}
\toprule
\multirow{2}{*}{\textbf{Model}} & \multirow{2}{*}{\textbf{\#Params (M)}} & \multirow{2}{*}{\textbf{CPU ms}}
  & \multicolumn{3}{c}{\textbf{Thai}}
  & \multicolumn{3}{c}{\textbf{Myanmar$^\dagger$}}
  & \multicolumn{3}{c}{\textbf{Malay$^\dagger$}}
  & \multicolumn{3}{c}{\textbf{Indonesian}}
  & \multicolumn{3}{c}{\textbf{Javanese$^\dagger$}}
  & \multicolumn{3}{c}{\textbf{Sundanese$^\dagger$}}
  & \multicolumn{3}{c}{\textbf{Vietnamese}}
  & \multicolumn{3}{c}{\textbf{Avg}} \\
\cmidrule(lr){4-6}\cmidrule(lr){7-9}\cmidrule(lr){10-12}\cmidrule(lr){13-15}\cmidrule(lr){16-18}\cmidrule(lr){19-21}\cmidrule(lr){22-24}\cmidrule(lr){25-27}
  & & & @1 & @5 & @10 & @1 & @5 & @10 & @1 & @5 & @10 & @1 & @5 & @10 & @1 & @5 & @10 & @1 & @5 & @10 & @1 & @5 & @10 & @1 & @5 & @10 \\
\midrule
Teacher & $86.19+499.77$ & \textit{90.5}
  & \textit{46.7} & \textit{73.8} & \textit{82.9}
  & \textit{23.4} & \textit{47.0} & \textit{58.9}
  & \textit{54.9} & \textit{81.0} & \textit{88.5}
  & \textit{58.5} & \textit{82.6} & \textit{90.4}
  & \textit{26.7} & \textit{50.9} & \textit{63.4}
  & \textit{24.4} & \textit{47.1} & \textit{58.6}
  & \textit{55.0} & \textit{81.4} & \textit{89.5}
  & \textit{41.4} & \textit{66.3} & \textit{76.0} \\
\midrule
CLIP-KD$^{\ast}$ & $5.62+40.49$ & 24.1
  & 0.1 & 0.6 & 1.0
  & 0.1 & 0.8 & 1.7
  & 1.5 & 4.2 & 6.2
  & 2.1 & 5.5 & 7.9
  & 2.2 & 7.3 & 10.9
  & 3.2 & 8.4 & 13.4
  & 0.2 & 0.6 & 1.1
  & 1.3 & 3.9 & 6.0 \\
TinyCLIP & $8.28+15.17$ & 14.9
  & 0.0 & 0.4 & 1.1
  & 0.3 & 1.4 & 1.8
  & 1.5 & 4.5 & 7.1
  & 2.2 & 6.3 & 9.1
  & 2.6 & 6.8 & 10.8
  & 3.3 & 9.1 & 14.3
  & 0.9 & 2.3 & 3.8
  & 1.5 & 4.4 & 6.9 \\
MobileCLIP2 & $11.41+63.43$ & 61.6
  & 0.6 & 1.3 & 1.8
  & 0.8 & 2.4 & 3.5
  & 17.7 & 35.1 & 44.8
  & \textbf{25.8} & 46.6 & 56.9
  & \textbf{10.4} & \textbf{25.3} & 33.9
  & \textbf{12.2} & \textbf{25.9} & \textbf{35.6}
  & 2.3 & 5.3 & 6.8
  & 10.0 & 20.3 & 26.2 \\
\midrule
\textbf{SEA-CLIP-Tiny} & $5.62+40.49$ & 24.1
  & \textbf{8.0} & \textbf{22.6} & \textbf{32.4}
  & \textbf{3.3} & \textbf{11.0} & \textbf{17.7}
  & \textbf{21.6} & \textbf{46.7} & \textbf{60.2}
  & 21.3 & \textbf{48.0} & \textbf{61.3}
  & 8.9 & 24.6 & \textbf{35.1}
  & 6.6 & 19.9 & 28.2
  & \textbf{15.4} & \textbf{37.6} & \textbf{50.2}
  & \textbf{12.2} & \textbf{30.0} & \textbf{40.7} \\
\bottomrule\end{tabular}
}
\caption{Per-language performance (mean across Babel-IN = acc@k, XM3600, Flickr30k-200, XTD-200 T2I (text$\to$image) R@$k$), reported as \%. \textbf{Bold} = best among student models per column. $\dagger$~Averaged over Flickr30k-200 and XTD-200 only; XM3600 does not cover these languages. $*$~Retrained by on CC12M following CLIPKD's scripts.}
\vspace{-8mm}
\label{tab:per_language_rk_t2i}
\end{table}
\vspace{-8mm}
\begin{table}[h!]\centering\setlength{\tabcolsep}{3pt}
\resizebox{\linewidth}{!}{%
\begin{tabular}{l c c ccc ccc ccc ccc ccc c}
\toprule
\multirow{3}{*}{\textbf{Model}} & \multirow{3}{*}{\textbf{\#Params (M)}} & \multirow{3}{*}{\textbf{CPU ms}} & \multicolumn{12}{c}{\textbf{Retrieval} (R@$k$)} & \multicolumn{3}{c}{\textbf{Classification} (acc@$k$)} & \textbf{VQA} \\
\cmidrule(lr){4-15}\cmidrule(lr){16-18}\cmidrule(lr){19-19}
  & & & \multicolumn{3}{c}{XM3600} & \multicolumn{3}{c}{Flickr30k-200} & \multicolumn{3}{c}{XTD-200} & \multicolumn{3}{c}{\textbf{Avg}} & \multicolumn{3}{c}{Babel-IN} & CVQA \\
\cmidrule(lr){4-6}\cmidrule(lr){7-9}\cmidrule(lr){10-12}\cmidrule(lr){13-15}\cmidrule(lr){16-18}
  & & & @1 & @5 & @10 & @1 & @5 & @10 & @1 & @5 & @10 & @1 & @5 & @10 & @1 & @5 & @10 & acc@1 \\
\midrule
Teacher & $86.19+499.77$ & \textit{90.5}
  & \textit{52.6} & \textit{78.5} & \textit{86.8} & \textit{45.0} & \textit{68.0} & \textit{76.1} & \textit{38.1} & \textit{64.8} & \textit{76.4} & \textit{45.2} & \textit{70.5} & \textit{79.7} & \textit{42.3} & \textit{61.3} & \textit{67.8} & \textit{48.6} \\
\midrule
CLIP-KD$^{\ast}$ & $5.62+40.49$ & 24.1
  & 0.7 & 1.8 & 2.5 & 1.1 & 2.9 & 4.4 & 1.7 & 5.1 & 8.0 & 1.1 & 3.3 & 5.0 & 4.2 & 7.7 & 10.5 & 25.9 \\
TinyCLIP & $8.28+15.17$ & 14.9
  & 1.0 & 2.8 & 3.8 & 1.5 & 3.6 & 5.6 & 1.6 & 5.2 & 8.6 & 1.4 & 3.9 & 6.0 & 4.4 & 8.2 & 10.8 & 26.8 \\
MobileCLIP2 & $11.41+63.43$ & 61.6
  & 9.2 & 17.3 & 20.8 & 9.2 & 18.0 & 23.2 & 10.9 & 22.7 & 29.6 & 9.8 & 19.3 & 24.5 & 15.6 & 23.5 & 27.5 & \textbf{33.5} \\
\midrule
SEA-CLIP-Tiny & $5.62+40.49$ & 24.1
  & \textbf{15.4} & \textbf{36.2} & \textbf{47.1} & \textbf{10.7} & \textbf{25.1} & \textbf{33.9} & \textbf{13.4} & \textbf{34.9} & \textbf{47.9} & \textbf{13.2} & \textbf{32.1} & \textbf{43.0} & \textbf{17.2} & \textbf{33.3} & \textbf{41.1} & 28.2 \\
\bottomrule\end{tabular}
}
\caption{Per-task SEA performance, reported as \%.
  Babel-IN = acc@$k$ (SEA-lang avg); XM3600, Flickr30k-200, XTD-200 = T2I (text$\to$image) R@$k$;
  CVQA = acc@1 (LOCAL, SEA-7 subset).
  \textbf{Retrieval Avg} = mean(Babel-IN, XM3600, Flickr30k-200, XTD-200).
  \textbf{Total Avg} = equal-weight mean over all 5 tasks.
  \textbf{Bold} = best student per column.}
  \vspace{-8mm}
\label{tab:per_task_rk_t2i}
\end{table}

\subsection{Human Preference Score} \label{appendix:human}

\begin{figure}[h!]
    \centering
    \vspace{-5mm}
    \includegraphics[width=\linewidth]{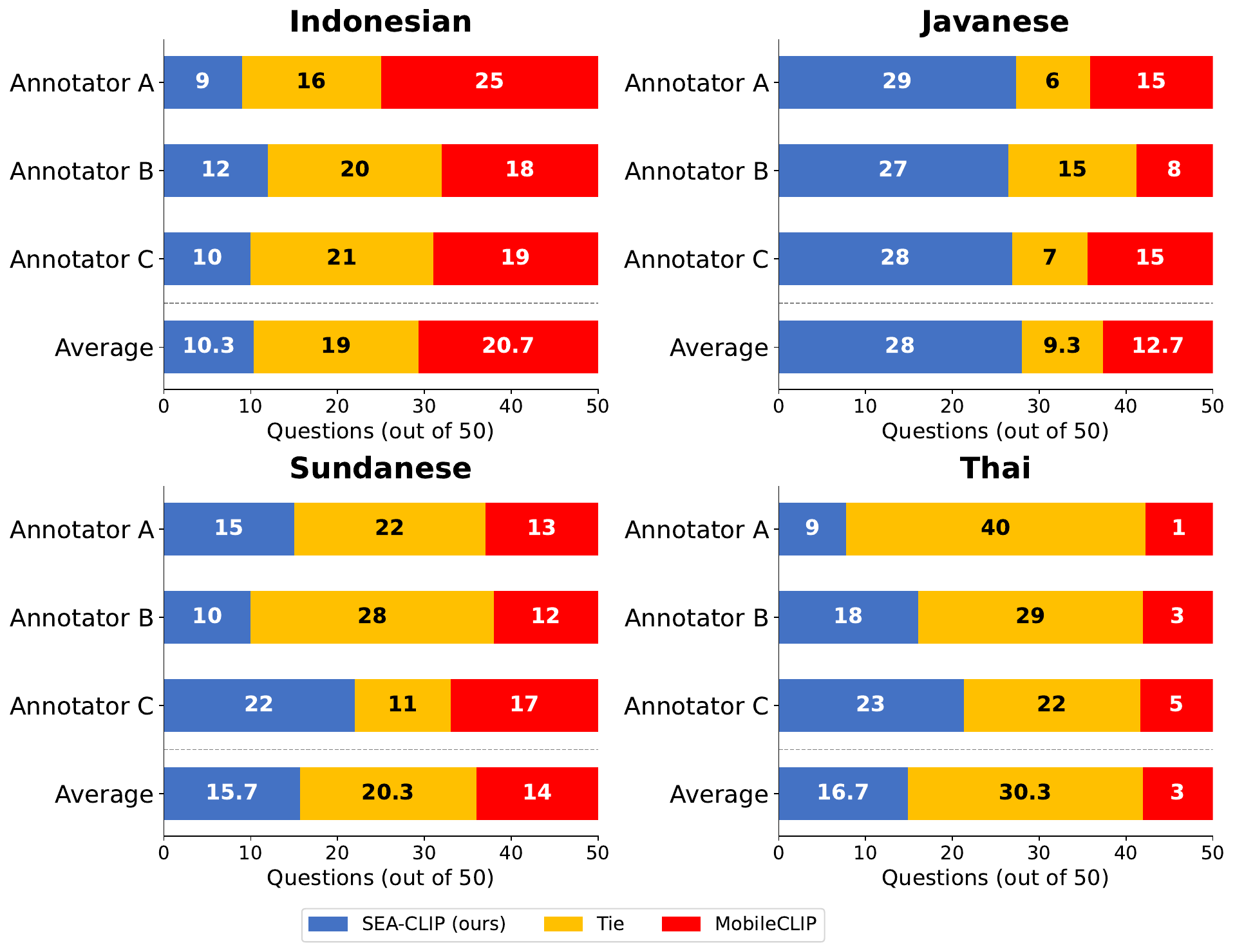}
    \caption{Human preference evaluation on XTD-200 only. Each language contains 50 samples. Bars show the number of questions where SEA-CLIP-Tiny wins, ties with MobileCLIP2, or MobileCLIP2 wins, reported per annotator and on average.}
    \vspace{-3mm}
    \label{fig:he_xtd}
\end{figure}

\begin{figure}[h!]
    \centering
    \vspace{-5mm}
    \includegraphics[width=\linewidth]{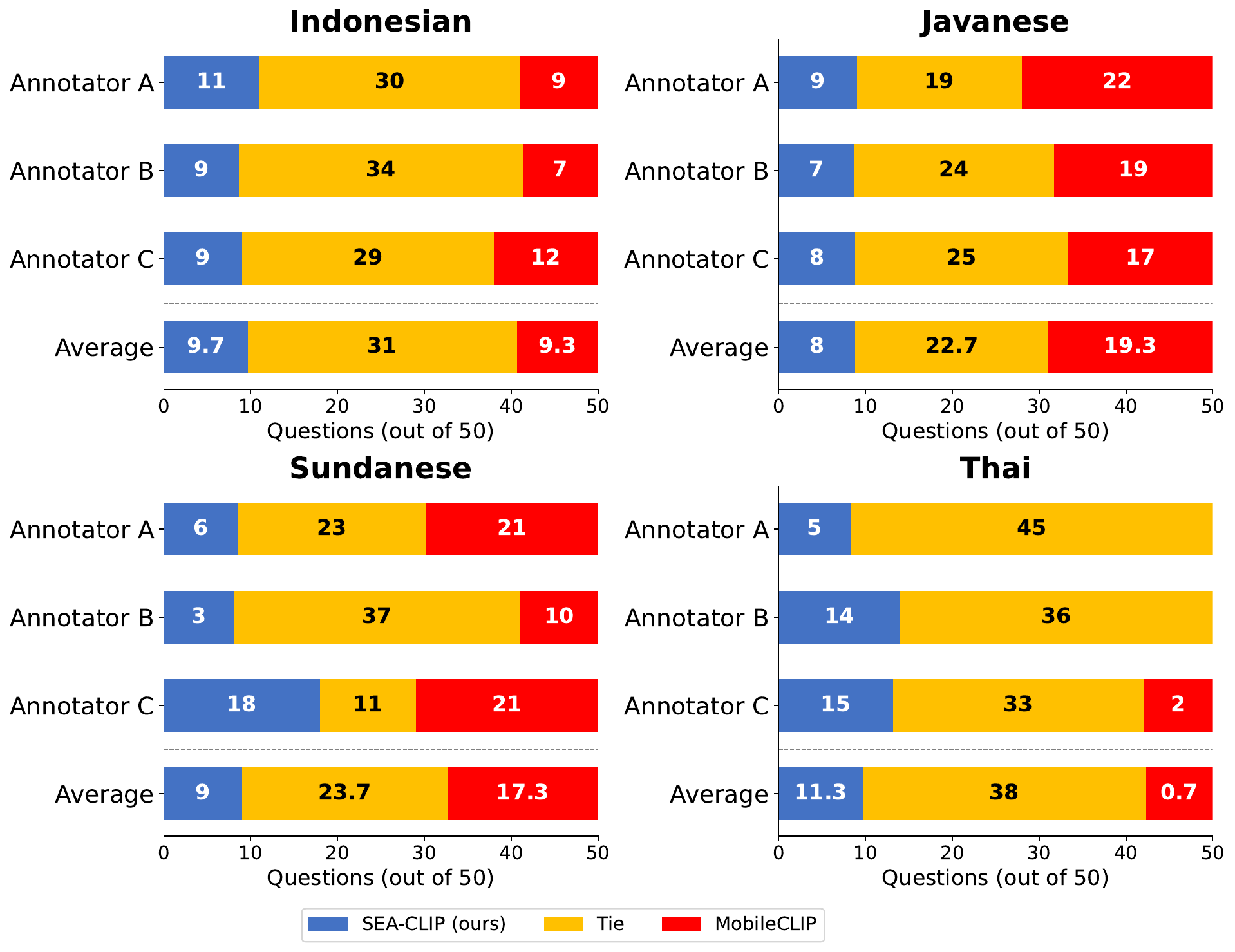}
    \caption{Human preference evaluation on Flickr30k-200 only. Each language contains 50 samples. Bars show the number of questions where SEA-CLIP-Tiny wins, ties with MobileCLIP2, or MobileCLIP2 wins, reported per annotator and on average.}
    \vspace{-3mm}
    \label{fig:he_flickr}
\end{figure}

\noindent
\textbf{Discussion.}
We further break down the human preference results by dataset in Figures~\ref{fig:he_xtd} and \ref{fig:he_flickr}. The results show similar trends to the main human evaluation. On Flickr30k-200, many examples are judged as ties, especially for Thai, Indonesian, and Javanese, suggesting that both models often produce similarly acceptable retrieval outputs. SEA-CLIP-Tiny is clearly preferred for Thai, while MobileCLIP2 is stronger for Sundanese. On XTD-200, SEA-CLIP-Tiny performs more favorably, winning clearly on Thai and Sundanese and remaining close to MobileCLIP2 on Javanese. These results suggest that the perceived gap between SEA-CLIP-Tiny and MobileCLIP2 is dataset-dependent, but SEA-CLIP-Tiny remains competitive from the perspective of native speakers, especially for Thai and Javanese.

\begin{figure}[h!]
    \centering
    
    \includegraphics[width=0.5\linewidth]{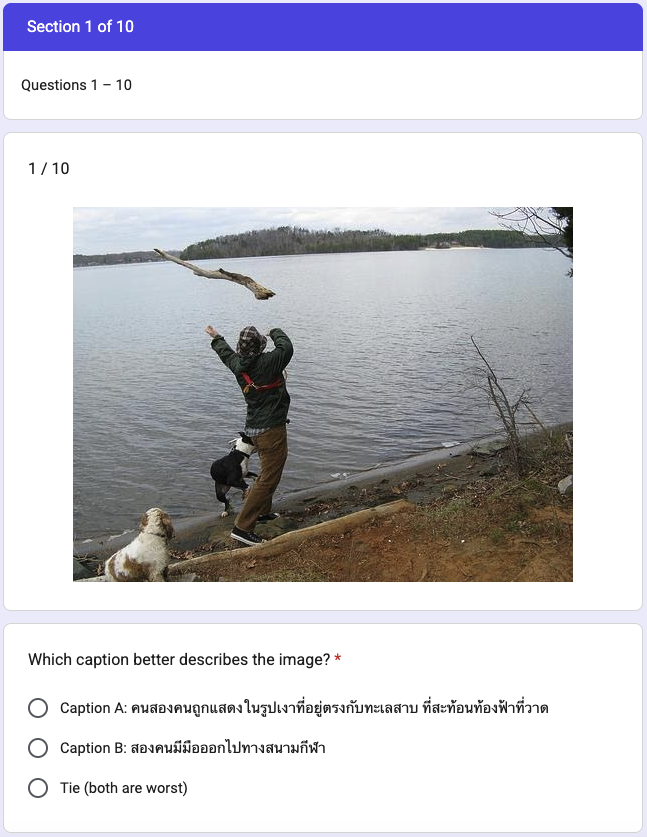}
    \caption{Example interface for the human preference evaluation. Annotators compare two anonymized captions for the same image and select the better caption, or mark a tie when both captions are equally acceptable or equally incorrect.}
    \vspace{-3mm}
    \label{fig:inference}
\end{figure}

\noindent
\textbf{Guidelines.} For the human preference guideline, annotators are asked to compare the top retrieved captions produced by the two models for the same image query. The model identities are hidden and shown only as Caption A and Caption B (Figure~\ref{fig:inference}). Annotators select the caption that better describes the image in the target language, considering both semantic correctness and naturalness for native speakers. If both captions are equally correct or equally incorrect, annotators are instructed to select \textit{Tie}. This setup allows us to measure which model retrieves captions that are more appropriate from the perspective of native speakers.

\noindent
\textbf{Details.} For each language, we recruit three annotators. All annotators are volunteer undergraduate students from Southeast Asia and are native speakers of the language they evaluate. They are also able to understand English, which allows them to follow the annotation instructions while judging the quality and naturalness of captions in their native language.

\begin{table}[h!]
\centering
\setlength{\tabcolsep}{4pt}
\resizebox{\linewidth}{!}{%
\begin{tabular}{l l l l ccccccc r}
\toprule
\multirow{2}{*}{\textbf{Benchmark}}
& \multirow{2}{*}{\textbf{Task}}
& \multirow{2}{*}{\textbf{Metric}}
& \multirow{2}{*}{\textbf{Provenance}}
& \multicolumn{7}{c}{\textbf{SEA language coverage}}
& \multirow{2}{*}{\textbf{$n$/lang}} \\
\cmidrule(lr){5-11}
& & & & Tha & Mya & Mal & Ind & Jav & Sun & Vie & \\
\midrule
Babel-ImageNet
& Zero-shot classification
& acc@$k$
& BabelNet lexical mapping
& \checkmark & \checkmark & \checkmark & \checkmark & \checkmark & \checkmark & \checkmark
& 98--896 \\

XM3600
& Image--text retrieval
& R@$k$
& Native-speaker annotation
& \checkmark & -- & -- & \checkmark & -- & -- & \checkmark
& 3,600 \\

Flickr30k-200
& Image--text retrieval
& R@$k$
& Machine translation (NLLB)
& \checkmark & \checkmark & \checkmark & \checkmark & \checkmark & \checkmark & \checkmark
& 1,000 \\

XTD-200
& Image--text retrieval
& R@$k$
& Machine translation (NLLB)
& \checkmark & \checkmark & \checkmark & \checkmark & \checkmark & \checkmark & \checkmark
& 1,000 \\

\midrule
CVQA
& Visual question answering
& acc@1
& Native-speaker authored
& \multicolumn{7}{c}{\textit{no per-language split released}}
& -- \\
\bottomrule
\end{tabular}
}
\caption{
Coverage and provenance of the evaluation benchmarks.
XM3600 covers Thai, Indonesian, and Vietnamese; therefore, the per-language retrieval averages for these languages in Table~\ref{tab:per_language} are computed over three benchmarks, while Malay, Javanese, Sundanese, and Myanmar are averaged over Flickr30k-200 and XTD-200.
Flickr30k-200 and XTD-200 rely on machine-translated text rather than native annotation, which is an important limitation for evaluating lower-resource languages.
Babel-ImageNet also has uneven test-set sizes across languages.
Filipino, Khmer, and Lao occur in our training mixture but are not covered by these evaluation sets and are therefore excluded from the main language-level evaluation.
$n$/lang denotes the number of evaluation examples per language.
}
\label{tab:coverage}
\end{table}

\begin{table}[h!]
\centering
\caption{
Efficiency comparison on a single NVIDIA A100-SXM4-40GB.
GFLOPs are measured per sample for one forward pass using
\texttt{torch.utils.flop\_counter} (1 MAC = 2 FLOPs) at each model's native
input resolution and context length.
Latency is measured at batch size 1 after 20 warm-up iterations and 100 timed runs,
while throughput and peak memory are measured at batch size 1,024.
Parameters include the final projection heads and token-embedding tables.
}
\label{tab:efficiency}
\resizebox{\textwidth}{!}{%
\begin{tabular}{lrrrrrrrrr}
\toprule
& \multicolumn{2}{c}{Params (M)}
& \multicolumn{2}{c}{GFLOPs/sample}
& \multicolumn{2}{c}{Latency (ms, bs=1)}
& \multicolumn{2}{c}{Throughput (samples/s, bs=1024)}
& Peak mem. \\
\cmidrule(lr){2-3}
\cmidrule(lr){4-5}
\cmidrule(lr){6-7}
\cmidrule(lr){8-9}
Model
& Image & Text
& Image & Text
& Image & Text
& Image & Text
& (MB) \\
\midrule
TinyCLIP
& 8.28 & 15.17
& 3.57 & \textbf{0.38}
& \textbf{5.27} & \textbf{2.60}
& \textbf{14{,}479} & \textbf{112{,}589}
& 2{,}278 \\

MobileCLIP2-S0
& 11.41 & 63.43
& 4.79 & 3.88
& 14.44 & 5.14
& 1{,}930 & 14{,}364
& 9{,}258 \\

\textbf{SEA-CLIP-Tiny}
& \textbf{5.62} & 40.49
& \textbf{2.51} & 3.38
& 6.57 & 6.65
& 13{,}213 & 18{,}544
& \textbf{1{,}739} \\
\midrule
MetaCLIP-2 (ViT-B-16-worldwide@WorldWideCLIP) (teacher)
& 86.19 & 499.77
& 35.13 & 5.96
& 6.58 & 6.81
& 3{,}429 & 15{,}112
& 7{,}745 \\
\bottomrule
\end{tabular}}
\end{table}

\begin{table}[h!]
\centering
\caption{
Single-sample CPU inference latency measured in fp32 at batch size 1 on an
AMD EPYC 7713 with \texttt{OMP\_NUM\_THREADS=8}.
We report the median of 30 runs after 5 warm-up iterations.
FLOPs are reported using CUDA profiling in Table~\ref{tab:efficiency}, since the CPU fused-attention implementation is not fully captured by our FLOP counter.
}
\label{tab:efficiency-cpu}
\resizebox{\textwidth}{!}{%
\begin{tabular}{lrrr}
\toprule
Model & Image (ms) & Text (ms) & Image + text (ms) \\
\midrule
TinyCLIP
& 12.52 & \textbf{2.40} & \textbf{14.92} \\

MobileCLIP2-S0
& 41.57 & 20.03 & 61.60 \\

\textbf{SEA-CLIP-Tiny}
& \textbf{10.40} & 13.73 & 24.13 \\
\midrule
MetaCLIP-2 (ViT-B-16-worldwide@WorldWideCLIP) (teacher)
& 70.24 & 20.25 & 90.49 \\
\bottomrule
\end{tabular}}
\end{table}

\begin{table}[h!]
\centering
\caption{
Parameter breakdown of the text and image towers. We report the token-embedding table separately because published compact CLIP models use different parameter-counting conventions. TinyCLIP excludes the text token-embedding table from its reported text-encoder size; under its published convention, our implementation reproduces the reported parameter counts. For consistency, our efficiency comparisons include the embedding table and final
projection heads for all models.
}
\label{tab:efficiency-params}
\resizebox{\textwidth}{!}{%
\begin{tabular}{lrrrr}
\toprule
& \multicolumn{3}{c}{Text tower (M)}
& Image tower (M) \\
\cmidrule(lr){2-4}
\cmidrule(lr){5-5}
Model
& Embedding table
& Transformer + head
& Total
& Total \\
\midrule
TinyCLIP
& 12.65 & 2.52 & 15.17 & 8.28 \\

MobileCLIP2-S0
& 25.30 & 38.13 & 63.43 & 11.41 \\

\textbf{SEA-CLIP-Tiny}
& 18.97 & 21.52 & 40.49 & \textbf{5.62} \\
\midrule
MetaCLIP-2 (ViT-B-16-worldwide@WorldWideCLIP) (teacher)
& 461.63 & 38.13 & 499.77 & 86.19 \\
\bottomrule
\end{tabular}}
\end{table}

\begin{table}[h!]
\centering
\caption{
Tokenizer vocabulary coverage of Southeast Asian scripts.
For each tokenizer, we count vocabulary entries containing at least one character
from the corresponding Unicode script.
\textbf{VI Latin} counts entries containing Vietnamese-specific Latin characters
with diacritics.
\textbf{SEA-script total} is the union of Thai, Myanmar, Khmer, and Lao script
entries and therefore excludes Latin-script entries.
\emph{Byte frag.} denotes student vocabulary entries that do not decode to a valid
UTF-8 character independently.
}
\label{tab:tok-vocab}
\resizebox{\textwidth}{!}{%
\begin{tabular}{lrrrrrrrr}
\toprule
Tokenizer
& Vocab.
& Thai
& Myanmar
& Khmer
& Lao
& VI Latin
& SEA-script total
& Byte frag. \\
\midrule
Student CLIP-BPE
& 49{,}408
& 51
& 0
& 0
& 0
& 59
& 51
& 532 \\

Teacher XLM-V-Base
& 901{,}629
& 14{,}243
& 4{,}214
& 3{,}029
& 2{,}112
& 58{,}773
& 23{,}598
& -- \\
\bottomrule
\end{tabular}}
\end{table}

\begin{table}[h!]
\centering
\caption{
Student CLIP-BPE tokenization statistics by benchmark and language.
Token lengths are measured before truncation and include start- and end-of-text tokens.
$n$ denotes the number of evaluated text inputs.
\emph{P95} and \emph{Max} summarize the pre-truncation token-length distribution, and
\emph{Trunc.} is the percentage of inputs exceeding the context length of 77 tokens.
}
\label{tab:tok-bench}
\small
\resizebox{\textwidth}{!}{%
\begin{tabular}{llrrrrrrr}
\toprule
Benchmark & Language & $n$ & Chars/text & Mean & Median & P95 & Max & Trunc. (\%) \\
\midrule
XM3600
& English    & 3{,}600 & 51.9 & 13.0 & 13 & 20  & 31  & 0.0 \\
& Indonesian & 3{,}600 & 93.3 & 31.4 & 29 & 55  & 95  & 0.1 \\
& Vietnamese & 3{,}600 & 74.9 & 68.4 & 66 & 116 & 193 & 33.3 \\
& Thai       & 3{,}600 & 54.0 & 59.2 & 59 & 100 & 163 & 26.1 \\
\midrule
Flickr30k-200
& English    & 1{,}000 & 94.6  & 22.3  & 20  & 35  & 86  & 0.1 \\
& Indonesian & 1{,}000 & 106.0 & 37.1  & 34  & 61  & 141 & 1.5 \\
& Javanese   & 1{,}000 & 96.3  & 34.4  & 32  & 56  & 135 & 1.0 \\
& Malay      & 1{,}000 & 107.4 & 38.0  & 35  & 61  & 136 & 1.7 \\
& Sundanese  & 1{,}000 & 98.8  & 36.6  & 34  & 62  & 148 & 1.3 \\
& Vietnamese & 1{,}000 & 102.5 & 91.7  & 86  & 155 & 310 & 60.7 \\
& Thai       & 1{,}000 & 72.4  & 79.1  & 72  & 142 & 287 & 42.6 \\
& Myanmar    & 1{,}000 & 102.5 & 195.0 & 182 & 337 & 642 & 95.7 \\
\midrule
XTD-200
& English    & 1{,}000 & 64.8 & 15.8  & 15  & 22  & 48  & 0.0 \\
& Indonesian & 1{,}000 & 69.1 & 24.1  & 23  & 36  & 87  & 0.2 \\
& Javanese   & 1{,}000 & 63.5 & 22.3  & 21  & 33  & 85  & 0.1 \\
& Malay      & 1{,}000 & 71.4 & 25.1  & 24  & 37  & 84  & 0.2 \\
& Sundanese  & 1{,}000 & 66.0 & 23.5  & 22  & 35  & 85  & 0.2 \\
& Vietnamese & 1{,}000 & 71.0 & 66.3  & 63  & 101 & 213 & 22.6 \\
& Thai       & 1{,}000 & 46.2 & 50.9  & 48  & 79  & 251 & 5.7 \\
& Myanmar    & 1{,}000 & 69.8 & 132.1 & 124 & 201 & 524 & 94.8 \\
\midrule
Babel-IN
& English    & 1{,}000 & 10.0 & 4.2  & 4  & 7  & 10 & 0.0 \\
& Indonesian & 463     & 10.3 & 5.5  & 5  & 8  & 12 & 0.0 \\
& Javanese   & 183     & 9.2  & 5.4  & 5  & 8  & 10 & 0.0 \\
& Malay      & 419     & 11.7 & 5.9  & 6  & 9  & 15 & 0.0 \\
& Sundanese  & 98      & 6.9  & 5.1  & 5  & 7  & 32 & 0.0 \\
& Vietnamese & 523     & 10.3 & 10.8 & 10 & 19 & 27 & 0.0 \\
& Thai       & 896     & 11.0 & 14.0 & 12 & 30 & 41 & 0.0 \\
& Myanmar    & 232     & 9.3  & 20.6 & 20 & 36 & 44 & 0.0 \\
\bottomrule
\end{tabular}}
\end{table}

\begin{table}[h!]
\centering
\caption{Caption statistics per language, macro-averaged over the three retrieval
benchmarks (XM3600, Flickr30k-200, XTD-200) at a context length of 77 tokens.
All \emph{Trunc}, \emph{Frag} and \emph{unk} values are percentages.
\emph{Trunc} is the share of captions exceeding the context length;
\emph{Frag} is the share of emitted student tokens that are UTF-8 byte fragments;
\emph{unk} is the share of emitted teacher tokens that are the unknown token.}
\label{tab:tokenizer_stats}
\footnotesize
\setlength{\tabcolsep}{4pt}
\renewcommand{\arraystretch}{1.1}
\begin{tabular}{lrrrrrrrr}
\toprule
Language
& \multicolumn{4}{c}{Student CLIP-BPE}
& \multicolumn{4}{c}{Teacher XLM-V-Base} \\
\cmidrule(lr){2-5}\cmidrule(lr){6-9}
& Tok/cap & Tok/char & Trunc & Frag
& Tok/cap & Tok/char & Trunc & unk \\
\midrule
English    & 17.0  & 0.21 & 0.0  & 0.0  & 17.3 & 0.25 & 0.0 & 0.00 \\
Indonesian & 30.9  & 0.32 & 0.6  & 0.0  & 17.2 & 0.19 & 0.0 & 0.00 \\
Javanese   & 28.4  & 0.33 & 0.6  & 0.0  & 18.9 & 0.24 & 0.1 & 0.00 \\
Malay      & 31.6  & 0.33 & 0.9  & 0.0  & 17.4 & 0.20 & 0.0 & 0.00 \\
Sundanese  & 30.1  & 0.34 & 0.8  & 0.0  & 20.9 & 0.25 & 0.1 & 0.00 \\
Vietnamese & 75.5  & 0.89 & 38.9 & 50.7 & 21.9 & 0.26 & 0.1 & 0.15 \\
Thai       & 63.1  & 1.06 & 24.8 & 55.9 & 15.1 & 0.26 & 0.0 & 0.00 \\
Burmese    & 163.6 & 1.87 & 95.2 & 99.8 & 28.3 & 0.33 & 0.8 & 0.00 \\
\bottomrule
\end{tabular}
\end{table}
\end{document}